\documentclass{article}

\usepackage{iclr2022_conference,times}

\usepackage{amsmath,amsfonts,bm}

\def\eqref#1{equation~\ref{#1}}

\def\1{\bm{1}}

\DeclareMathAlphabet{\mathsfit}{\encodingdefault}{\sfdefault}{m}{sl}
\SetMathAlphabet{\mathsfit}{bold}{\encodingdefault}{\sfdefault}{bx}{n}

\usepackage[utf8]{inputenc} 
\usepackage[T1]{fontenc}    
\usepackage{hyperref}       
\usepackage{url}            
\usepackage{booktabs}       
\usepackage{amsfonts}       
\usepackage{nicefrac}       
\usepackage{microtype}      
\usepackage{xcolor}         

\usepackage[ruled,vlined,linesnumbered, resetcount]{algorithm2e}
\usepackage{graphicx}

\usepackage{subcaption}
\usepackage{multirow}
\usepackage[normalem]{ulem}
\useunder{\uline}{\ul}{}
\usepackage{textcomp}
\usepackage{footnote}
\usepackage{makecell}
\usepackage{xspace}
\usepackage{enumitem}
\usepackage{mdwlist}
\usepackage{wrapfig,lipsum,booktabs}
\usepackage{amsthm}

\newcommand{\prob}{\textsc{Lbrp}\xspace} 
\newcommand{\probf}{\textsc{Bfrp}\xspace} 
\newcommand{\cvnt}{\textsc{COVID-19}\xspace} 
\newcommand{\dtst}{\textsl{CoVDS}\xspace} 
\newcommand{\model}{\textsc{B2F}\xspace} 
\newcommand{\graphgen}{\textsc{GraphGen}\xspace} 
\newcommand{\bseqenc}{\textsc{BseqEnc}\xspace} 
\newcommand{\modelenc}{\textsc{ModelPredEnc}\xspace} 
\newcommand{\reiner}{\textsc{Refiner}\xspace} 
\newcommand{\bseq}{\textsc{BSeq}\xspace} \newcommand{\bseqs}{\textsc{BSeq}s\xspace} 
\newcommand{\revmae}{\textsc{RevDiffMAE}\xspace} 
\newcommand{\ensemble}{\textsc{Ensemble}\xspace} 
\newcommand{\deepcovid}{\textsc{GT-DC}\xspace} 
\newcommand{\yyg}{\textsc{YYG}\xspace} 
\newcommand{\umass}{\textsc{UMass-MB}\xspace} 
\newcommand{\cmu}{\textsc{CMU-TS}\xspace} 

\newcommand{\bseqreg}{\textsc{BseqReg}\xspace}
\newcommand{\bseqregt}{\textsc{BseqReg2}\xspace}
\newcommand{\modelreg}{\textsc{PredRNN}\xspace} 
\newcommand{\ffnreg}{\textsc{FFNReg}\xspace} 

\newcommand{\rregions}{\mathrm{Reg}} 
\newcommand{\ffeat}{\mathrm{Feat}} 
\newcommand{\fftall}{\mathcal{F}} 
\newcommand{\frr}{\texttt{RetailRec}\xspace} 
\newcommand{\fgc}{\texttt{Grocery}\xspace} 
\newcommand{\fpk}{\texttt{Parks}\xspace}
\newcommand{\ftr}{\texttt{Transit}\xspace} 
\newcommand{\fwk}{\texttt{WorkSpace}\xspace} 
\newcommand{\frs}{\texttt{Resident}\xspace}
\newcommand{\fcvn}{\texttt{HospRate}\xspace} 
\newcommand{\fpinc}{\texttt{+veInc}\xspace} 
\newcommand{\fninc}{\texttt{-veInc}\xspace}
\newcommand{\ftinc}{\texttt{TestResultsInc}\xspace} 
\newcommand{\fvnt}{\texttt{onVentilator}\xspace} 
\newcommand{\ficu}{\texttt{inICU}\xspace} 
\newcommand{\frec}{\texttt{Recovered}\xspace} 
\newcommand{\fhosp}{\texttt{HospInc}\xspace} 
\newcommand{\fdt}{\texttt{Deaths}\xspace} 
\newcommand{\fdxa}{\texttt{DexA}\xspace} 
\newcommand{\fapm}{\texttt{AppleMob}\xspace} 
\newcommand{\ffac}{\texttt{Facilities}\xspace} 
\newcommand{\fvs}{\texttt{ERVisits}\xspace} 
\newcommand{\fbcli}{\texttt{FbCLI}\xspace} 
\newcommand{\fbwili}{\texttt{FbWiLi}\xspace} 

\newcommand{\bed}{\textsf{Early Decrease}\xspace} 
\newcommand{\beinc}{\textsf{Early Increase}\xspace} 
\newcommand{\bst}{\textsf{Steady/Spike}\xspace} 
\newcommand{\blin}{\textsf{Late Increase}\xspace} 
\newcommand{\bmd}{\textsf{Mid Decrease}\xspace} 

\newtheorem{obs}{Obs.}
\newtheorem{defn}{Defn.}

\newcommand{\hub}{COVID-19 Forecast Hub\xspace}

\newcommand{\be}{\textsc{BErr}\xspace}

\newcommand{\bd}{\textsc{STime}\xspace}

\title{Back2Future: Leveraging Backfill Dynamics for Improving Real-time Predictions in Future}

\author{%
  Harshavardhan Kamarthi, Alexander Rodr\'iguez, B. Aditya Prakash\\
  College of Computing\\
  Georgia Institute of Technology\\
  \texttt{\{harsha.pk,arodriguezc,badityap\}@gatech.edu} \\
}

\iclrfinalcopy
\begin{document}

\maketitle
\begin{abstract}

For real-time forecasting in domains like public health and macroeconomics, data collection is a non-trivial and demanding task. Often after being initially released, it undergoes several revisions later (maybe due to human or technical constraints) - as a result, it may take weeks until the data reaches a stable value. This so-called ‘backfill’ phenomenon and its effect on model performance have been barely addressed in the prior literature. In this paper, we introduce the multi-variate backfill problem using COVID-19 as the motivating example. 
We construct a detailed dataset composed of relevant signals over the past year of the pandemic. 
We then systematically characterize several patterns in backfill dynamics and leverage our observations for formulating a novel problem and neural framework, Back2Future, that aims to refines a given model's predictions in real-time. Our extensive experiments demonstrate that our method refines the performance of diverse set of top models for COVID-19 forecasting and GDP growth forecasting. Specifically, we show that Back2Future refined top COVID-19 models by 6.65\% to 11.24\% and yield 18\% improvement over non-trivial baselines. In addition, we show that our model improves model evaluation too; hence policy-makers can better understand the true accuracy of forecasting models in real-time.
\end{abstract}

\section{Introduction}

The current COVID-19 pandemic has challenged our response capabilities to large disruptive events, affecting the health and economy of millions of people. 
A major tool in our response has been forecasting epidemic trajectories, which has provided lead time to policymakers to optimize and plan interventions~\citep{holmdahl_wrong_2020}.
Broadly two classes of approaches  have been devised: traditional mechanistic epidemiological models~\citep{shaman2012forecasting,zhang2017forecasting}, and the fairly newer statistical approaches~\citep{brooks_nonmechanistic_2018,adhikari2019epideep,osthus2019dynamic} including deep learning models~\citep{adhikari2019epideep,deng2020cola, panagopoulos2020transfer,rodriguez_steering_2021}, which have become among the top-performing ones for multiple forecasting tasks~\citep{reich_collaborative_2019}. These also leverage newer digital indicators like search queries~\citep{ginsberg2009detecting,yang2015accurate} and social media~\citep{culotta2010towards,lampos2010flu}. As noted in multiple previous works~\citep{metcalf2017opportunities,biggerstaff2018results}, epidemic forecasting is a challenging enterprise because it is affected by weather, mobility, strains, and others. 

However, \emph{real-time} forecasting also brings new challenges. 
As noted in multiple CDC real-time forecasting initiatives for diseases like flu~\citep{osthus2019even} 
and COVID-19~\citep{cramer2021evaluation}, as well as in macroeconomics \citep{clements2019data,aguiar2015macroeconomic} 
the initially released public health data is revised many times after and is known as the 'backfill' phenomenon.
The various factors that affect backfill are multiple and complex, ranging from surveillance resources to human factors like coordination between health institutes and government organizations within and across regions \citep{chakraborty_what_2018,reich_collaborative_2019,Altieri2021Curating,stierholz2017economic}.

While previous works have addressed anomalies~\citep{liu2017holoscope}, missing data~\citep{yin2020identifying}, and data delays \citep{vzliobaite2010change} in general time-series problems, the backfill problem has not been addressed.
In contrast, the topic of revisions has not received as much attention, with few exceptions. 
For example in epidemic forecasting, a few papers have either (a) mentioned about the `backfill problem' and its effects on performance~\citep{chakraborty_what_2018,rodriguez_deepcovid_2021,Altieri2021Curating,rangarajan2019forecasting} and evaluation~\citep{reich_collaborative_2019}; or (b) proposed to address the problem via simple models like linear regression~\citep{chakraborty2014forecasting} or 'backcasting'~\citep{brooks_nonmechanistic_2018} the observed targets c) used data assimilation and sensor fusion from a readily available stable set of features to refine unrevised features \citep{farrow2016modeling,osthus2019even}. However, they focus only on revisions in the \emph{target} and typically study in the context of influenza forecasting, which is substantially less noisy and more regular than the novel COVID-19 pandemic or assume access to stable values for some features which is not the case for COVID-19.   
In economics, \cite{clements2019data} surveys several domain-specific~\citep{carriero2015forecasting} or essentially linear techniques for data revision/correction behavior of  several macroeconomic indicators~\citep{croushore2011forecasting}.

Motivated from above, we study the more challenging problem of multi-variate backfill for both features and targets. We go further beyond prior work and also show how to leverage our insights for a more general neural framework to \emph{improve} both predictions (i.e. \emph{refinement} of the model's predictions) and performance evaluation (i.e. \emph{rectification} from the evaluator's perspective). 
Our specific contributions are the following: 


\noindent$\bullet$ \textbf{Multi-variate backfill problem:} We introduce the multi-variate backfill problem using real-time epidemiological forecasting as the primary motivating example. In this challenging setting, which generalizes (the limited) prior work, the forecast targets, as well as exogenous features, are subject to retrospective revision. Using a carefully collected diverse dataset for COVID-19 forecasting for the past year, we discover several patterns in backfill dynamics, show that there is a significant difference in real-time and revised feature measurements, and highlight the negative effects of using unrevised features for incidence forecasting in different models both for model performance and evaluation. Building on our empirical observations, we formulate the problem \probf, which aims to ‘correct’ given model predictions to achieve better performance on eventual fully revised data.

\noindent$\bullet$ \textbf{Spatial and Feature level backfill modeling to refine model predictions:} Motivated by the patterns in revision and observations from our empirical study, we propose a deep-learning model Back2Future (\model) to model backfill revision patterns and derive latent encodings for features. 
\model combines Graph Convolutional Networks that capture sparse, cross-feature, and cross-regional backfill dynamics similarity and deep sequential models that capture temporal dynamics of each features' backfill dynamics across time.
The latent representation of all features is used along with the history of the model's predictions to improve diverse classes of models trained on real-time targets, to predict targets closer to revised ground truth values.
Our technique can be used as a 
‘wrapper' to improve model performance of any forecasting model (mechanistic/statistical). 

\noindent$\bullet$ \textbf{ Refined top models' predictions and improved model evaluation:} We perform an extensive empirical evaluation to show that incorporating backfill dynamics through \model consistently improves the performance of diverse classes of top-performing COVID-19 forecasting models (from the CDC \hub, including the top-performing official ensemble) significantly.
We also utilize \model to help forecast evaluators and policy-makers better evaluate the  ‘eventual’ true accuracy of participating models (against revised ground truth, which may not be available until weeks later).  This allows the model evaluators to quickly estimate models that perform better w.r.t revised stable targets instead of potentially misleading current targets. Our methodology can be adapted for other time-series forecasting problems in general. We also show the generalizability of our framework and model \model to other domains by significantly improving predictions of non-trivial baselines for US National GDP forecasting \citep{marcellino2008linear,tkacz1999forecasting}.

\vspace{-7pt}
\section{Nature of backfill dynamics}
\vspace{-7pt}
\label{sec:obs}
In this section, we study important properties of the revision dynamics of our signals. We introduce some concepts and definitions to aid in the understanding of our empirical observations and method.


\par\noindent\textbf{Real-time forecasting.} 
We are given a set of signals $\fftall = \rregions \times \ffeat$, where $\rregions$ is the set of all regions (where we want to forecast) and set $\ffeat$ contains our features and forecasting target(s) for each region.
At prediction week $t$, $x^{(t)}_{i,1:t}$ is a time series from 1 to $t$ for feature $i$, and the set of all signals results in the multi-variate time series $\mathcal{X}^{(t)}_{1:t}$\footnote{In practice, delays are possible too, i.e, at week $t$, we have data for some feature $i$ only until $t-\delta_i$. All our results incorporate these situations. We defer the minor needed notational extensions to Appendix for clarity.}.
Similarly, $\mathcal{Y}^{(t)}_{1:t}$ is the forecasting target(s) time series.
Further, let's call all data available at time $t$, $\mathcal{D}^{(t)}_{1:t} = \{ \mathcal{X}^{(t)}_{1:t},  \mathcal{Y}^{(t)}_{1:t} \}$ as \emph{\underline{real-time sequence}}. 
For clarity we refer to `signal' $i \in \fftall $ as a sequence of either a feature or a target, and denote it as $d^{(t)}_{i,1:t}$. 
Thus, at prediction week $t$, the real-time forecasting problem is:
\emph{Given $\mathcal{D}^{(t)}_{1:t}$, predict next $k$ values of forecasting target(s), i.e. $\hat{y}_{t+1:t+k}$}. Typically for CDC settings and this paper, our time unit is week, $k=4$ (up to 4 weeks ahead) and our target is COVID-19  mortality incidence (\fdt).

\noindent\textbf{Revisions.} 
Data revisions (`backfill') are common. At prediction week $t+1$, the real-time sequence $\mathcal{D}^{(t+1)}_{1:t+1}$ is available. In addition to the length of the sequences increasing by one (new data point), values of $\mathcal{D}^{(t+1)}_{1:t+1}$ already in $\mathcal{D}^{(t)}_{1:t}$ may be revised i.e., $\mathcal{D}^{(t)}_{1:t} \neq \mathcal{D}^{(t+1)}_{1:t}$. 
Note that previous work has studied backfill limited to $\mathcal{Y}^{(t)}$, while we address it in both $\mathcal{X}^{(t)}$ and $\mathcal{Y}^{(t)}$.
Also, note that the data in the backfill is the same used for real-time forecasting, but just seen from a different perspective.

\noindent\textbf{Backfill sequences:} 
Another useful way we propose to look at backfill is by focusing on revisions of a single value. Let's focus on value of signal $i$ at an \emph{\underline{observation week}} $t'$. 
For this observation week, the value of the signal can be revised at any $t>t'$, which induces a sequence of revisions.
We refer to \emph{\underline{revision week}} $r\geq0$ as the relative amount of time that has passed since the observation week $t'$. 


\begin{defn}
(Backfill Sequence \bseq) For signal $i$ and observation week $t'$, its backfill sequence is $\bseq(i,t')= \langle d_{i,t'}^{(t')}, d_{i,t'}^{(t'+1)}, \ldots, d_{i,t'}^{(\infty)} \rangle$, where $d_{i,t'}^{(t')}$ is the \emph{\underline{initial value}} of the signal and $d_{i,t'}^{(\infty)}$ is the \emph{\underline{final/stable value}} of the signal.
\end{defn}
\begin{defn}
(Backfill Error \be) For revision week $r$ of a backfill sequence, the backfill error is
 $  \be(r,i,t') = | d^{(t'+r)}_{i,t'} - d^{(\infty)}_{i,t'}|~/~|d^{(\infty)}_{i,t'}|.$
\end{defn}
\begin{defn}
(Stability time \bd) of a backfill sequence \bseq is the revision week $r^*$ that is the minimum $r$ for which the backfill error \be  $< \epsilon$ for all $r>r^*$, i.e., the time when \bseq stabilizes. 
\end{defn}
\textit{Note:}  
We ensured that \bseq length is at least 7, and found that in our dataset most signals stabilize before $r=20$.
For  $d^{(\infty)}_{i,t'}$, we use $d^{(t_f)}_{i,t'}$, at the final week $t_f$ in our revisions dataset.
In case we do not find $\be<\epsilon$ in any \bseq, we set \bd to the length of that \bseq. We use $\epsilon=0.05$.
\textbf{Example:} For \bseq $\{223, 236, 236, 404, \dots, 404\}$,  \be for third week is $\frac{|236-404|}{404} = 0.41$ and \bd is 4.

\vspace{-7pt}
\subsection{Dataset description}
\vspace{-7pt}
\label{subsec:dataset}
\begin{wraptable}{r}{0.28\linewidth}
\vspace{-0.8cm}
    \centering
    \caption{List of features in our \dtst}
    \vspace{-2pt}
    \scalebox{0.8}{
    \begin{tabular}{p{1.6cm}|p{3cm}}
        \textbf{Type} &  \textbf{Features}\\ \hline
        Patient Line-List  & \fvs, \fcvn, \fpinc, \fhosp, \frec, \fvnt, \ficu\\ \hline 
        Testing  & \ftinc, \fninc, \ffac\\ \hline
        Mobility  & \frr, \fgc, \fpk, \ftr, \fwk, \frs, \fapm\\ \hline
        Exposure & \fdxa \\ \hline
        Social Survey  & \fbcli, \fbwili\\ \hline
    \end{tabular}
    }
    \label{tab:list_feats}
    \vspace{-.5cm}
\end{wraptable}
We collected important publicly available signals from a variety of trusted sources that are relevant to \cvnt forecasting to form the \textit{\cvnt Surveillance Dataset} (\dtst). 
See Table~\ref{tab:list_feats} for the list of 20 features ($|\ffeat| = 21$, including \fdt). 
Our revisions dataset contains signals that we collected every week since April 2020 and ends on July 2021.
Our analysis covers $30$ observation weeks from June 2020 to December 2020 
(to ensure all our backfill sequences are of length at least 7) 
for all $|\rregions| = 50$ US states. The rest of the \emph{unseen} data from Jan 2021 to July 2021 is used strictly for evaluation.
\looseness=-1

\par\textbf{Patient line-list:} traditional surveillance signals used in epidemiological models~\citep{chakraborty2014forecasting,brooks_nonmechanistic_2018} derived from line-list records e.g. hospitalizations from CDC~\citep{cdc_coronavirus_surveillance_2020}, positive cases, ICU admissions from COVID Tracking~\citep{covidtracking}. 
\textbf{Testing:} measure changes in testing from CDC and COVID-Tracking e.g. tested population, negative tests, used by \cite{rodriguez_deepcovid_2021}. \textbf{Mobility:} track people
movement to several point of interests (POIs), from \cite{google} and \cite{apple}, and serve as digital proxy for social distancing~\citep{arik2020interpretable}. 
\textbf{Exposure:} digital signal measuring closeness between people at  POIs, collected from mobile phones \citep{chevalier2021measuring} 
\textbf{Social Survey:} 
 previously used by \citep{wang2020covid,rodriguez_deepcovid_2021} CMU/Facebook Symptom Survey Data, which contains self-reported responses about COVID-19 symptoms.

\vspace{-7pt}
\subsection{Observations}
\vspace{-7pt}
We first study different facets of the significance of backfill in \dtst.
Using our definitions, we generate a backfill sequence for every combination of signal, observation week, and region (not all signals are available for all regions). In total, we generate \emph{more than} $30,000$ backfill sequences. 

\par\noindent\textbf{Backfill error \be is significant.}
We computed \be for the initial values, i.e., $\be(r=0,i,t')$, for all signals $i$ and observation weeks $t'$.

\begin{figure}[h]
    \hspace{-0.5cm}
    \centering
    \begin{tabular}{cccc}
    \centering
    \includegraphics[width=0.2\linewidth]{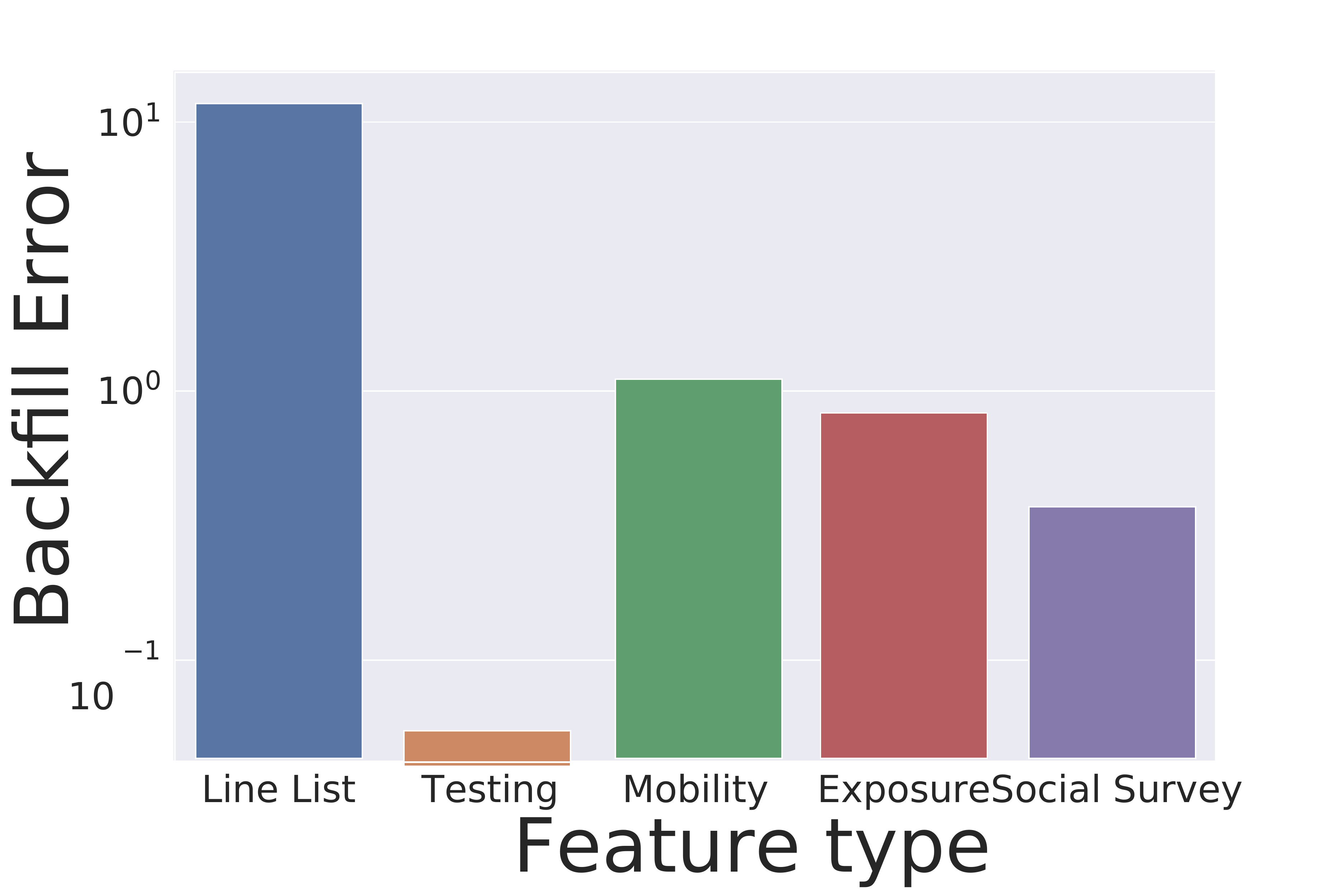} &
    \includegraphics[width=0.2\linewidth]{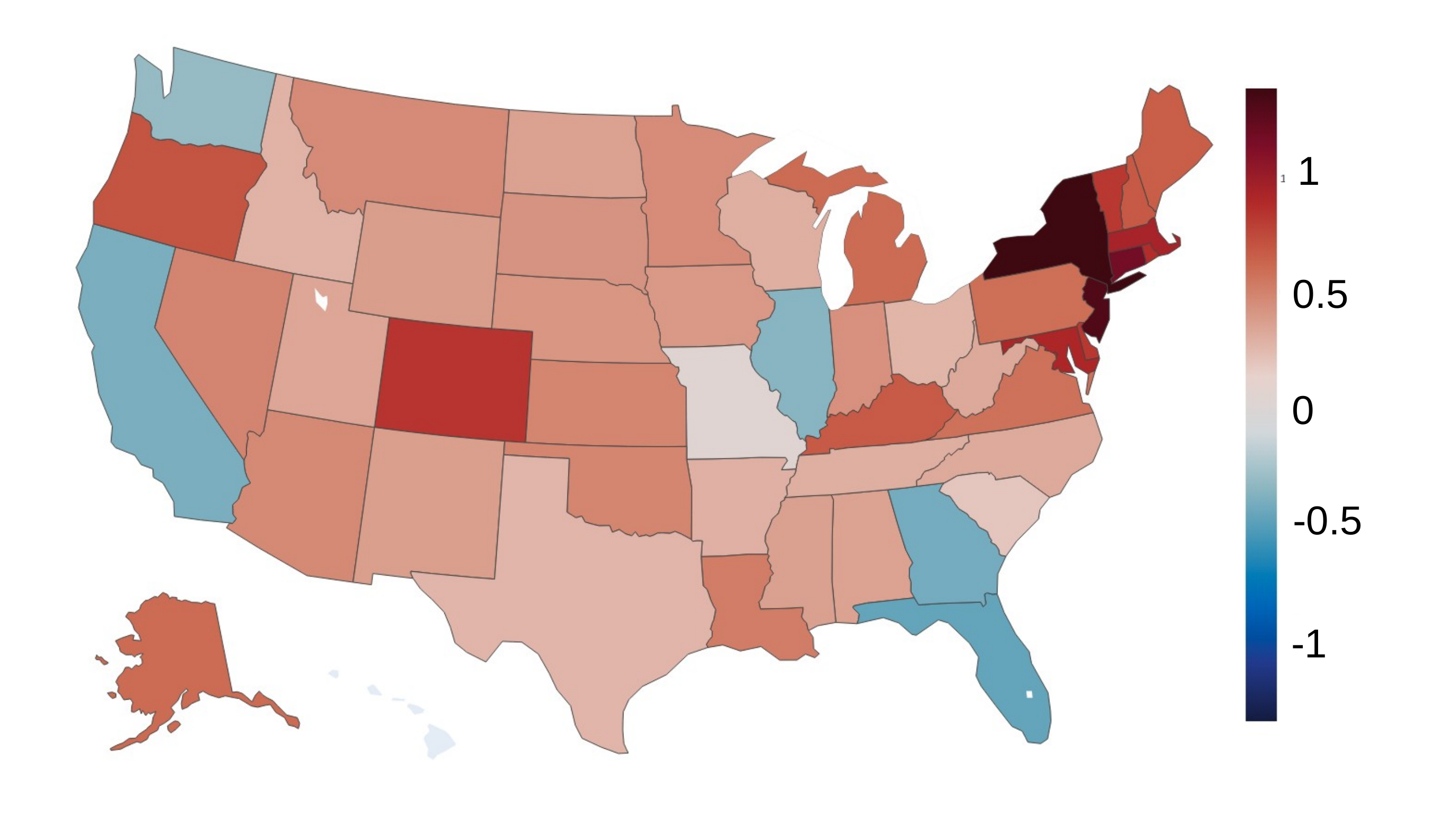} &
    \includegraphics[width=0.2\linewidth]{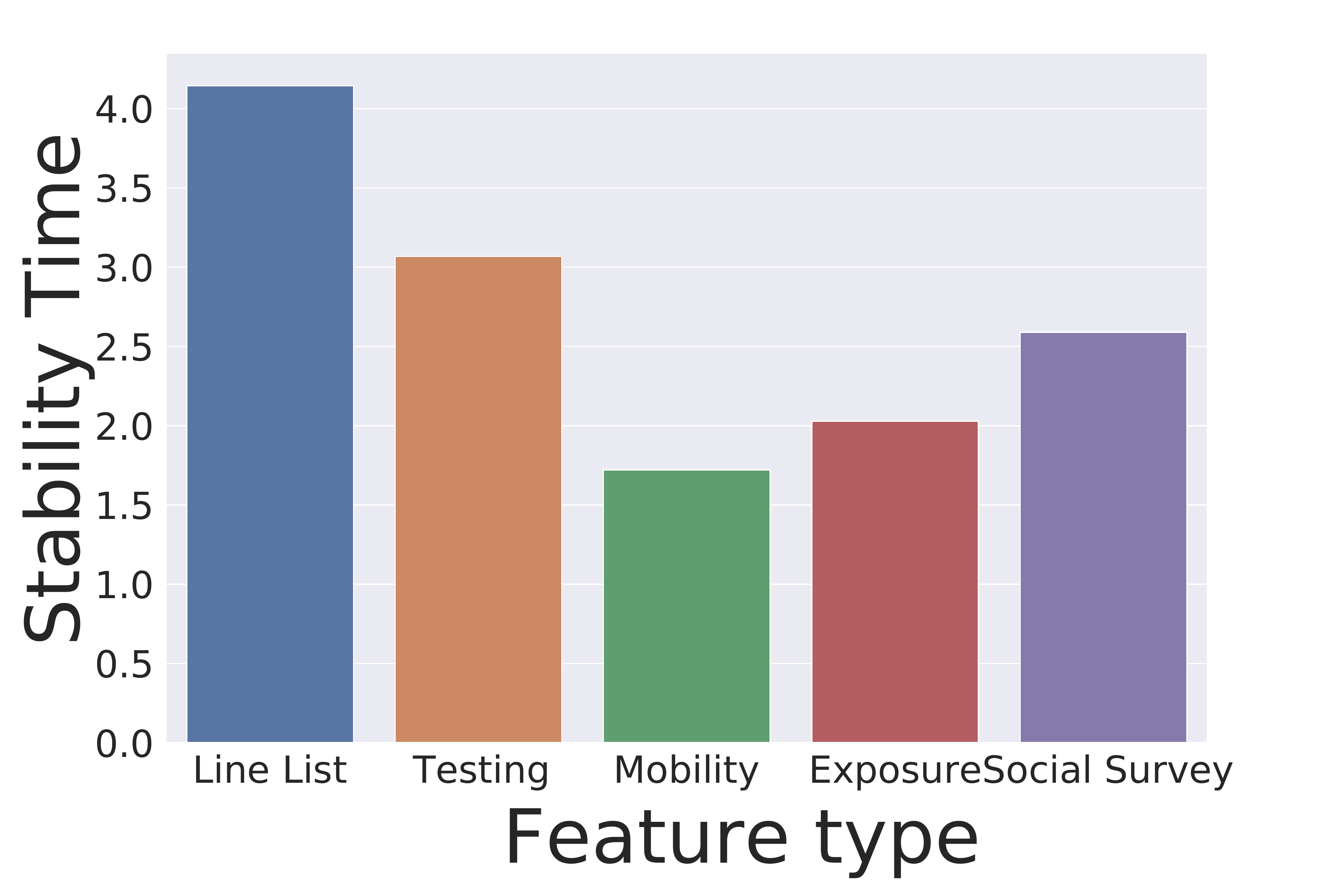} &
    \includegraphics[width=0.2\linewidth]{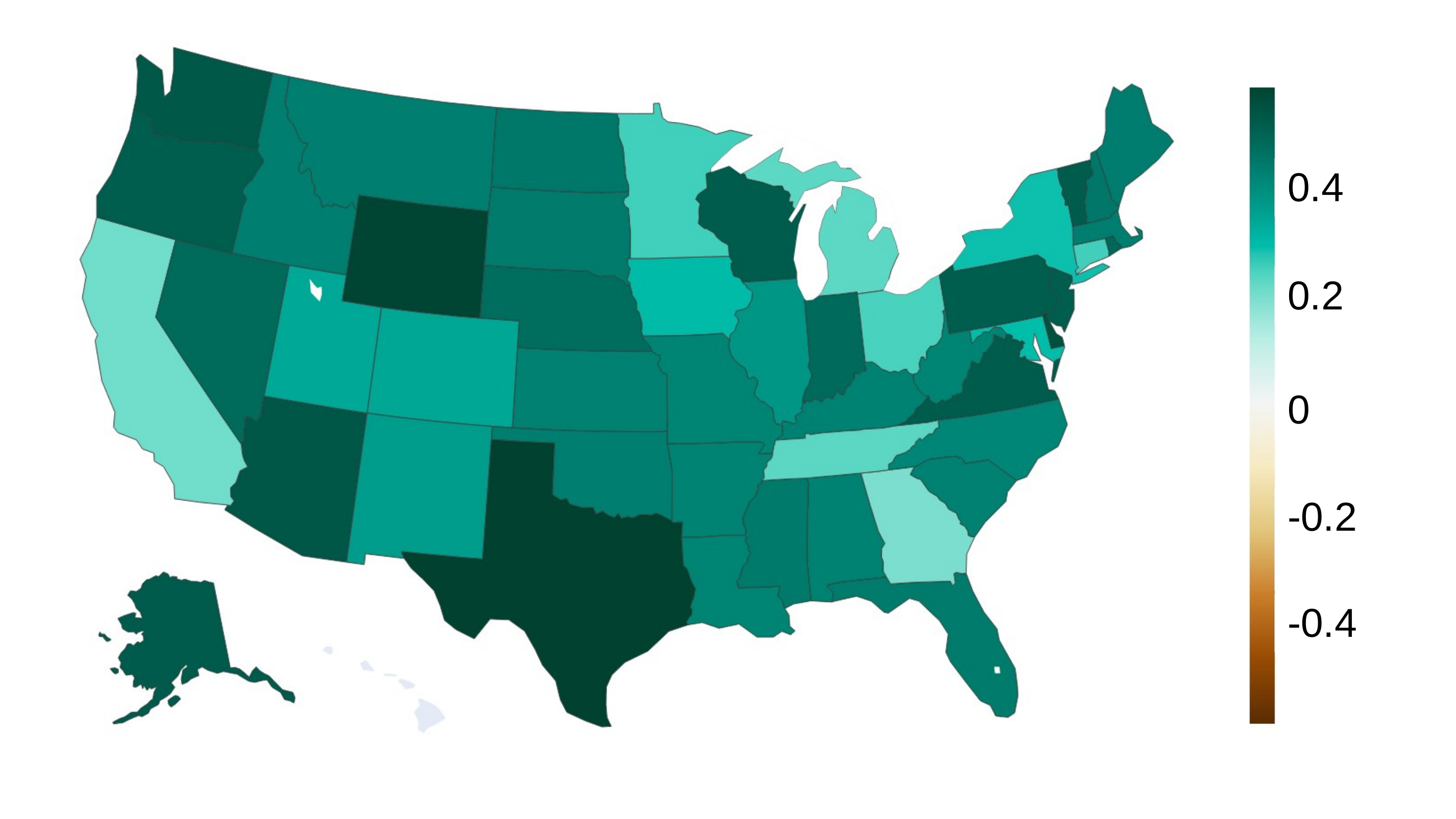} \\
    (a) \be per feat. type & (b) \be per region  & (c) \bd per feat. type & (d) \bd per region \\
  \end{tabular}
    \vspace{-4pt}
    \caption{\textit{\be and \bd across  \underline{feature type} and \underline{regions}, heat maps are $\log$ scaled.}}
    \label{fig:significant}
    \vspace{-0.08in}
\end{figure}

\begin{obs} (\be across signals and regions)
Compute the average \be for each signal; the median of all these averages is $32\%$, i.e. at least half of all signals are corrected by $32\%$ of their initial value. Similarly 
in at least half of the regions the signal corrections are 280\% of their initial value.
\label{obs:be}
\end{obs}
\vspace{-9pt}
We also found large variation of \be. 
For features (Figure~\ref{fig:significant}a), compare 
avg. $\be=1743\%$ of five most corrected features with $1.6\%$ of the five least corrected features.
Also, in contrast to related work that focuses on traditional surveillance data~\cite{yang2015accurate}, perhaps unexpectedly, we found that digital indicators also have a significant \be (average of $108\%$). 
For regions (see Figure~\ref{fig:significant}b), 
compare $1594\%$ of the five most corrected regions with $38\%$ of the five least corrected regions.



\vspace{-.1in}


\par\noindent\textbf{Stability time \bd is significant.}
A similar analysis for \bd found 
significant variation across signals (from 1 weeks for  to 21 weeks for \cvnt, see Figure~\ref{fig:significant}c for \bd across feature types) and regions (from 1.55 weeks for GA to 3.83 weeks for TX, see Figure~\ref{fig:significant}d). This also impacts our target, thus, actual accuracy is not readily available which undermines real-time evaluation and decision making.

\begin{obs} (\bd of features and target)
Compute the average \bd for each signal; the average of all these averages for features is around 4 weeks and for our target \fdt is around 3 weeks, i.e. on average, it takes over 3 weeks to reach the stable values of features.
\label{obs:bd}
\end{obs}
\vspace{-9pt}

\par
\begin{figure}[h]
    \centering
    \includegraphics[width=.9\linewidth,height=0.7in]{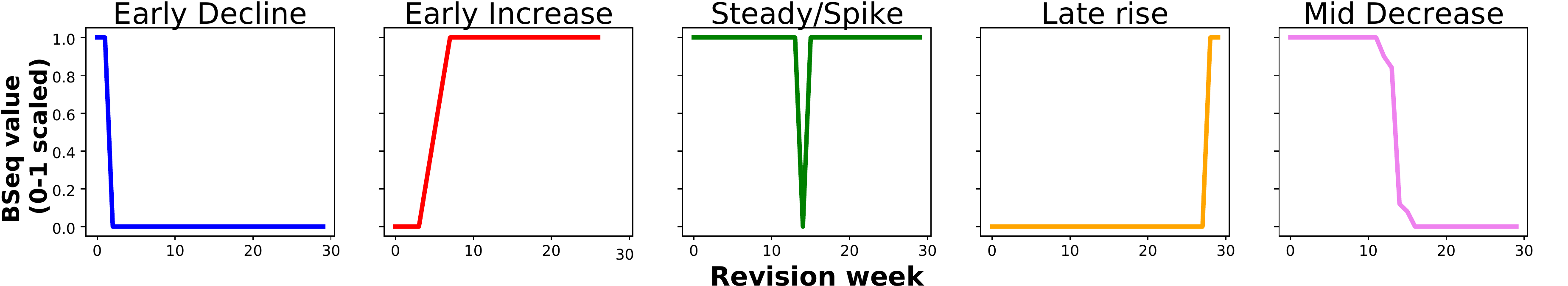}
    \vspace{-5pt}
    \caption{\textit{Centroid \bseq of each cluster, scaled between $[0,1]$, showing canonical backfill behaviors}
    }
    \vspace{-5pt}
    \label{fig:patterns}
\end{figure}

\noindent\textbf{Backfill sequence \bseq patterns.}
There is significant similarity among \bseqs.
We cluster \bseqs via K-means using Dynamic Time Warping (DTW) as pair-wise distance (as DTW can handle sequences of varying magnitude and length). We found five \emph{canonical} categories of behaviors (see Figure~\ref{fig:patterns}), each of size roughly 11.58\% of all \bseqs.
\vspace{-.05in}
\vspace{-.05in}
Also, each cluster is not defined only by signal nor region. Hence there is a non-trivial similarity across both signals and regions. 



\begin{obs} (\bseq similarity and variety)
Five canonical behaviors were observed in our backfill sequences (Figure~\ref{fig:patterns}). 
No cluster has over 21\% of \bseqs from the same region, and no cluster has over 14\% of \bseqs from the same signal.
\label{obs:patterns}
\end{obs}
\vspace{-9pt}

\begin{wrapfigure}{r}{.43\linewidth}
    \vspace{-0.3in}
    \centering
    \begin{tabular}{cccc}
    \centering
    \includegraphics[width=0.48\linewidth]{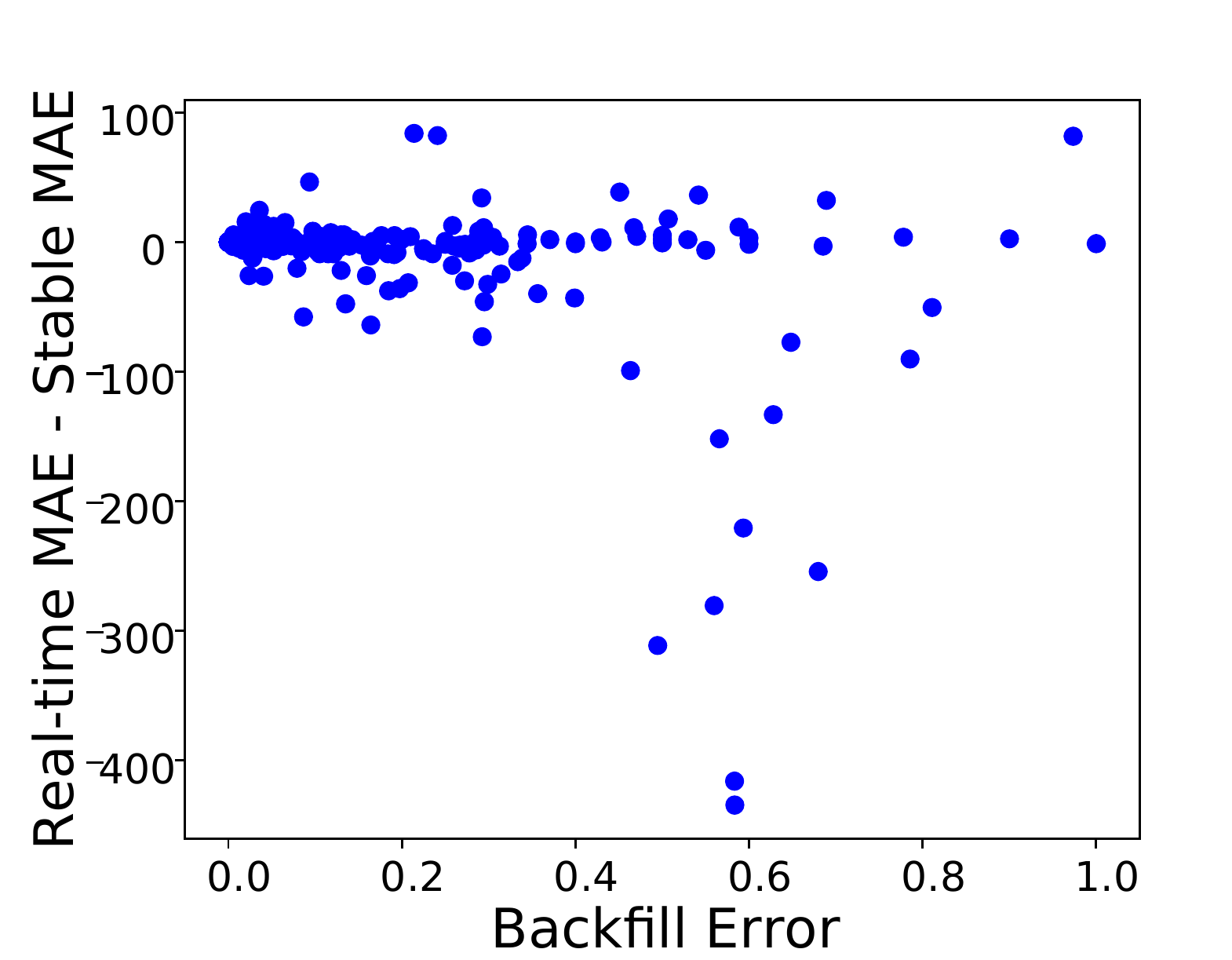} &
    \hspace{-10pt}
    \includegraphics[width=0.48\linewidth]{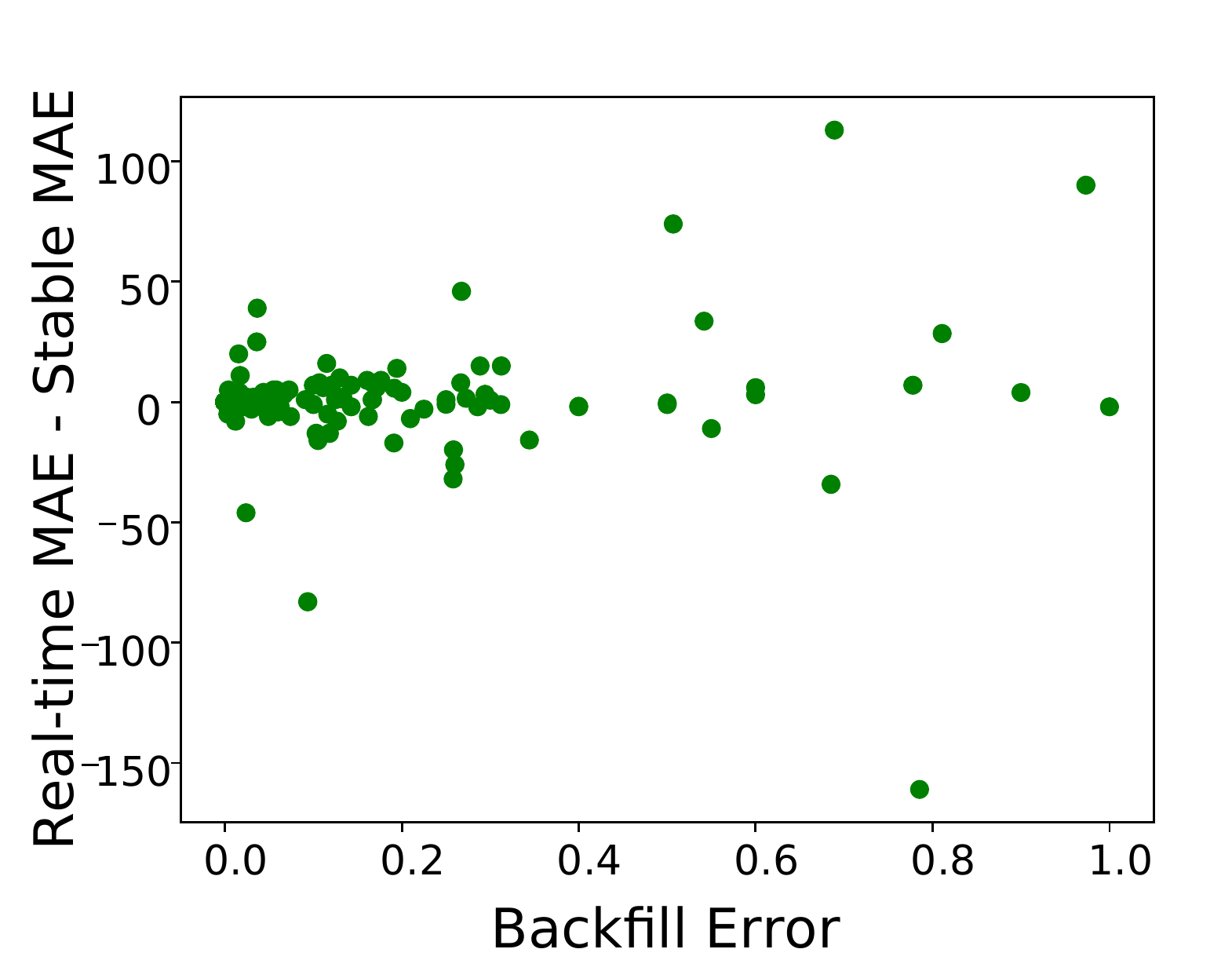}  \\
    (a) \deepcovid & (b) \yyg \\
  \end{tabular}
    \vspace{-0.15in}
    \caption{\textit{\be vs model \revmae.}}
    \label{fig:obs_corr}
    \vspace{-0.15in}
\end{wrapfigure}
\par\noindent\textbf{Model performance vs \be.}
To study the relationship between model performance (via Mean Absolute Error MAE of a prediction) and \be, we use \revmae: the difference between MAE computed against \emph{real-time} target value and one against the \emph{stable} target value. 
We analyze the top-performing real-time forecasting models as per the comprehensive evaluation of all models in \hub~\citep{cramer2021evaluation}. \yyg and \umass are mechanistic while \cmu and \deepcovid are statistical models.
The top performing \ensemble is composed of all contributing models to the hub.
We expect a well-trained real-time model will have higher \revmae with larger \be in its target \citep{reich_collaborative_2019}.
However, we found that higher \be does not necessarily mean worse performance. 
See Figure~\ref{fig:obs_corr}---\yyg has even better performance with more revisions. This may be due to the more complex backfill activity/dependencies in COVID in comparison to the more regular seasonal flu. 
\looseness=-1

\begin{wrapfigure}[7]{r}{.26\linewidth}
    \vspace{-0.3in}
    \centering
    \includegraphics[width=.9\linewidth]{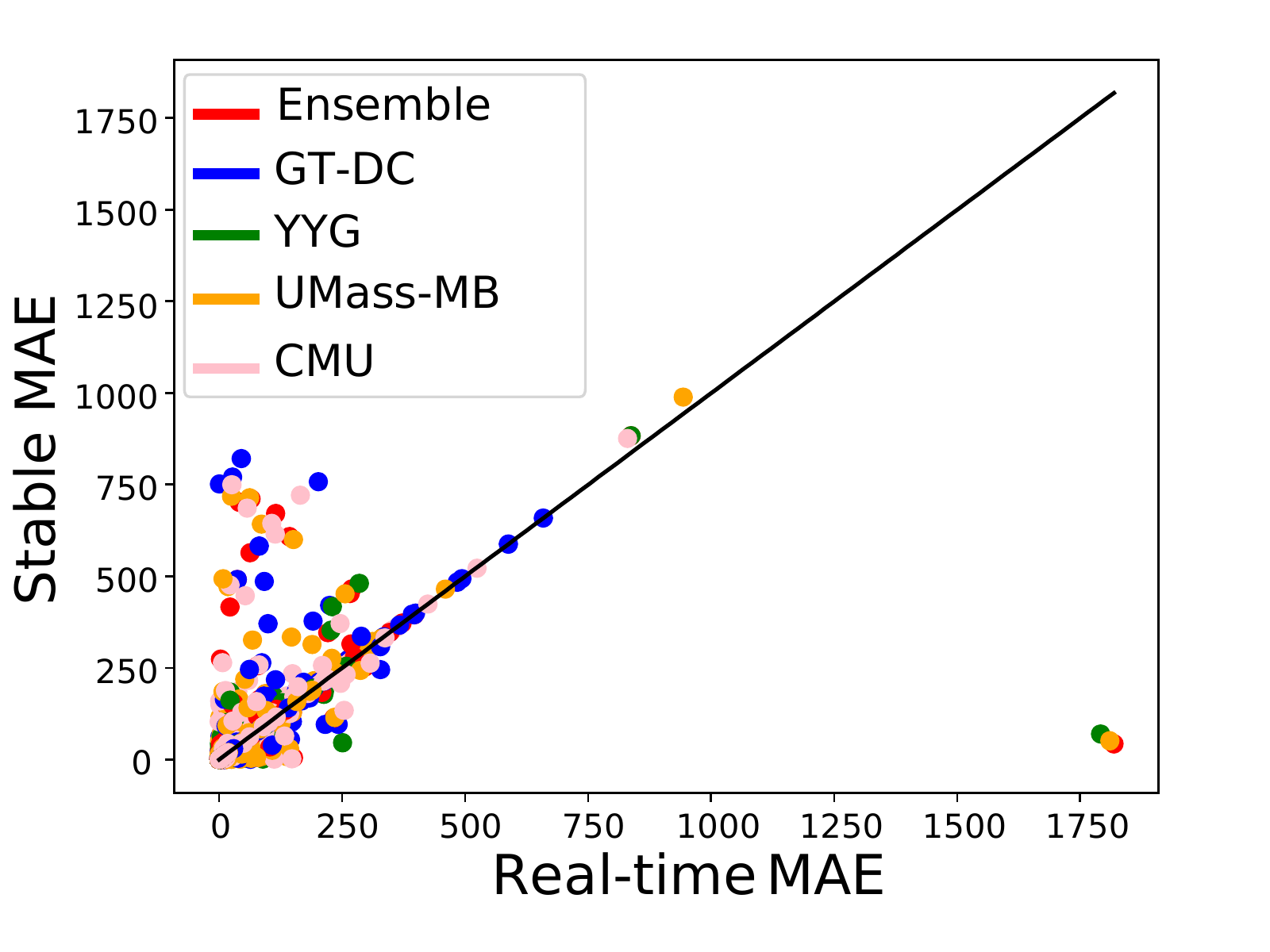}
    \label{fig:obs4_mae}
    \vspace{-0.13in}
    \caption{\textit{Real-time vs stable MAE.} 
    }
    \label{fig:obs4}
\end{wrapfigure}

\begin{obs} (Model performance and backfill)
Relation between \be and \revmae can be non-monotonous and positively or negatively correlated depending on model and signal. 
\label{obs:model}
\end{obs}
\noindent \textbf{Real-time target values to measure model performance}:
Since targets undergo revisions (5\% \be on average), we study how this \be affects the real-time evaluation of models. 
From Figure~\ref{fig:obs4}, we see that the scores are not similar with real-time scores over-estimating model accuracy. The average difference in scores is positive
which implies that evaluators would overestimate models' forecasting ability.

\begin{obs}
MAE evaluated at real-time overestimates model performance by 9.6 on average, with the maximum for TX at 22.63. 
\label{obs:eval}
\end{obs}
\vspace{-8pt}
\section{Refining the future via \model}
\vspace{-7pt}
Our observations naturally motivate improving training and evaluation aspects of \emph{real-time} forecasting by leveraging revision information. Thus, we propose the following two problems. 
Let predictions of model $M$ for week $t+k$ be $y(M,k)_t$. Since models are trained on \emph{real-time} targets, $y(M,k)_t$ is the  model's estimate of 
target $y_{t+k}^{(t+k)}$.

\textbf{$k$-week ahead Backfill Refinement Problem, $\probf_k(M)$:} At prediction week $t$, we are given 
a \emph{revision dataset}
$\{\mathcal{D}_{1:t'}^{(t')}\}_{t'\leq t}$, which includes our target $\mathcal{Y}^{(t)}_{1:t}$.
For a model $M$ trained on \emph{real-time} targets, given history of model's predictions till last week $\langle y(M,k)_1, \dots y(M,k)_{t-1}\rangle$ and prediction for current week $y(M,k)_{t}$, our goal is to refine $y(M,k)_{t}$ to better estimate the \emph{stable} target $y_{t+k}^{(t_f)}$, i.e. the 'future' of our target value at $t+k$.

\textbf{Leaderboad Refinement problem, \prob{}:} 
At each week $t$, evaluators are given a current estimate of our target $y_{t}^{(t)}$ and forecasts of models submitted on week $t-k$. Our goal is to refine $y_{t}^{(t)}$ to $\hat{y}_t$, a better estimate of $y_{t}^{(t_f)}$, so that using $\hat{y}_t$ as a surrogate for $y_{t}^{(t_f)}$ to evaluate predictions of models provides a better indicator of their actual performance (i.e., we obtain a refined leaderboard of models).
Relating it to \probf, assume a hypothetical model $M_{eval}$ whose predictions are real-time ground truth, i.e. $y(M_{eval},0)_t = y_{t}^{(t)}, \forall t$. Then, refining $M_{eval}$ is equivalent to refining $y_{t}^{(t)}$ to better estimate $y_{t}^{(t_f)}$ which leads to solving \prob{}. Thus, \emph{\prob{} is a \underline{special case} of $\probf_{0}(M_{eval})$}.
\par \noindent \textbf{Overview:} 
We leverage observations from Section~\ref{sec:obs} to derive Back2Future (\model), a deep-learning model that uses revision information from \bseq to refine predictions. Obs.~\ref{obs:be} and \ref{obs:bd} show that real-time values of signals are poor estimates of stable values. Therefore, we leverage patterns in \bseq of past signals and exploit cross-signal similarities (Obs. \ref{obs:patterns}) to extract information from \bseqs. We also consider that the relation of models' forecasts to \be of targets is complex (Obs. \ref{obs:model} and \ref{obs:eval}) to refine their predictions.
\model combines these ideas through its four modules:
    \noindent$\bullet$ \graphgen: Generates a signal graph (where each node maps to a signal in $\rregions \times \ffeat$) whose edges are based on  \bseq similarities.
     \noindent$\bullet$  \bseqenc: Leverages the signal graph as well as temporal dynamics of \bseqs to learn a latent representation of \bseqs using a Recurrent Graph Neural Network.
     \noindent$\bullet$  \modelenc: Encodes the history of the model's predictions, the real-time value of the target, and past revisions of the target through a recurrent neural network.
    \noindent$\bullet$ \reiner: Combines encodings from \bseqenc and \modelenc to predict the correction to model's real-time prediction.

\begin{wrapfigure}[14]{l}{.7\linewidth}
\centering
\vspace{-0.15in}
\includegraphics[width=\linewidth]{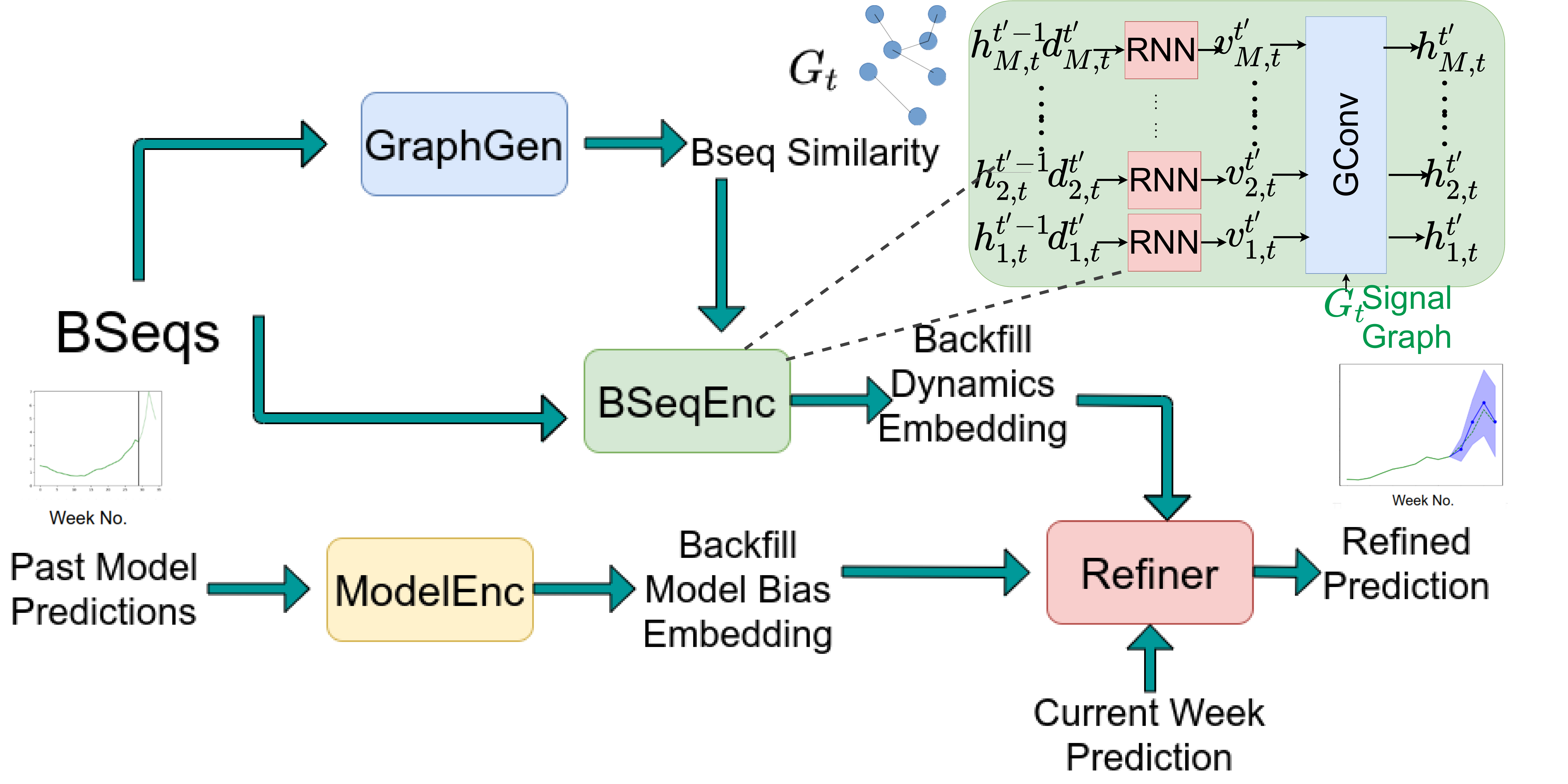}
\vspace{-0.28in}
\caption{\textit{\model pipeline with all components}}
\label{fig:bpipe}
\end{wrapfigure}

In contrast to previous works that studies target \be \citep{reich_collaborative_2019}, we simultaneously model all \bseq available till current week $t$ using spatial and signal similarities in the temporal dynamics of \bseq. Recent works that attempt to model spatial relations for COVID19 forecasting need explicitly structural data (like cross-region mobility) \citep{Panagopoulos2020UnitedWS} to generate a graph or use attention over temporal patterns of regions' death trends. 
\model, in contrast, directly models the structural information of signal graph (containing features from each region) using \bseq similarities. 
Thus, we first generate useful latent representations for each signal based on \bseq revision information of that feature as well as features that have shown similar revision patterns in the past. Due to the large number of signals that cover all regions, we cannot model the relations between every pair using fully connected modules or attention similar to \citep{Jin2020InterSeriesAM}. Therefore, we first construct a sparse graph between signals based on past \bseq similarities.
Then we inject this similarity information using Graph Convolutional Networks (GCNs) and combine it with deep sequential models to model temporal dynamics of \bseq of each signal while combining information from \bseq~s of signals in the neighborhood of the graph. Further, we use these latent representations and leverage the history of a model $M$'s predictions to refine its prediction.
Thus, \model solves $\probf_{k}(M)$ assuming $M$ is a black box, accessing only its past forecasts. Our training process, that involves pre-training on model-agnostic auxiliary task, greatly improves training time for refining any given model $M$. The full pipeline of \model is also shown in Figure \ref{fig:bpipe}.
Next, we describe each of the components of \model in detail. For the rest of this section, we will assume that we are forecasting $k$ weeks ahead given data till current week $t$. 
\vspace{-.1in}
\par \noindent \textbf{\graphgen} 
generates an undirected signal graph $G_t = (V, E_t)$ whose edges represent similarity in \bseqs between signals, where vertices $V = \fftall = \rregions \times \ffeat$. We measure similarity using DTW distance due to reasons described in Section \ref{sec:obs}. \graphgen leverages the similarities across \bseq patterns irrespective of the exact nature of canonical behaviors which may vary across domains. We compute the sum of DTW distances of \bseqs for each pair of nodes summed over $t'\in \{1,2,\dots, t-5\}$. We threshold $t'$ till $t-5$ to make the \bseqs to be of reasonable length (at least 5) to capture temporal similarity without discounting too many \bseqs. Top $\tau$ node pairs with lowest total DTW distance are assigned an edge. 

\par \noindent \textbf{\bseqenc.}
While we can model backfill sequences for each signal independently using a recurrent neural network, this doesn't capture the behavioral similarity of \bseq across signals. Using a fully-connected recurrent neural network that considers all possible interactions between signals also may not learn from the similarity information due to the sheer number of signals ($50\times 21=1050$) while greatly increasing the parameters of the model. Thus, we utilize the structural prior of graph $G_t$ generated by \graphgen and train an autoregressive model \bseqenc which consists of graph recurrent neural network to encode a latent representation for each of backfill sequence in $B_t = \{\bseq(i,t):i\in \fftall\}$. 
At week $t$, \bseqenc is first pre-trained 
and then it is fine-tuned 
for a specific model $M$ (more details later in this section). 

Our encoding process is in Figure \ref{fig:bpipe}. Let $
\bseq_{t'+r}(i,t')$ be first $r+1$ values of $\bseq(i,t')$ (till week $t'+r$).
For a past week $t'$ and revision week $r$, we denote $h_{i,t'}^{(t'_r)} \in \mathrm{R}^m$ to be the latent encoding of $\bseq_{t'_r}(i, t')$ where $t'_r = t'+r$ and $t'\leq t'_r\leq t$.
We initialize  $h_{i,t'}^{(0)}$ for any observation week $t'$ to be a learnable parameter $h_{i}^{(0)} \in \mathrm{R}^m$ specific to signal $i$. For each week $t'_r$ we combine latent encoding $h_{i,t'}^{(t'_r-1)}$ and signal value $d_{i,t'}^{(t'_r)}$ using a GRU (Gated Recurrent Unit)~\citep{Cho2014OnTP} cell to get the intermediate embedding $v_{i,t}^{(t')}$. Then, we leverage the signal graph $G_t$ and pass the embeddings $\{v_{i,t'}^{(t'_r)}: i\in \fftall\}$ through a Graph Convolutional layer \citep{kipf2016semi} to get $h_{i,t'}^{(t'_r)}$: \looseness=-3
{\small
\vspace{-0.07in}
\begin{equation}
v_{i,t'}^{(t'_r)} = \mathrm{GRU}_{\mathrm{BE}}(d_{i,t'}^{(t'_r)}, h_{i,t'}^{(t'_r-1)}), 
\quad \quad \quad 
\{h_{i,t'}^{(t'_r)}\}_{i\in \fftall} = \mathrm{GConv}(G_t, \{v_{i,t'}^{(t'_r)}\}_{i\in \fftall}).
\end{equation} 
}
\noindent Thus, $h_{i,t'}^{(t'_r)}$ contains information from $\bseq_{t'_r}(i,t')$ and structural priors from $G_t$.
Using $h_{i,t'}^{(t'_r)}$, \bseqenc predicts the value $d_{i,t'}^{(t'_r+1)}$ by passing through a 2-layer feed-forward network $\mathrm{FFN}_i$: $\hat{d}_{i,t'}^{(t'_r+1)} = \mathrm{FFN}_i(h_{i,t'}^{(t'_r)})$.
During inference, we only have access to real-time values of signals for the current week. We autoregressively predict $h_{i,t}^{(t+l)}$ for each signal by initially passing $\{d_{i,t}^{(t)}\}_{i\in \fftall}$ through \bseqenc and using the output $\{\hat{d}_{i,t}^{(t+1)}\}_{i\in \fftall}$ as input for \bseqenc. Iterating this $l$ times we get  $\{h_{i,t}^{(t+l)}\}_{i\in \fftall}$ along with $\{\hat{d}_{i,t}^{(t+l)}\}_{i\in \fftall}$ where $l$ is a hyperparameter.


\par \noindent \textbf{\modelenc.}
To learn from history of a model's predictions and its relation to target revisions, \modelenc encodes the history of model's predictions, previous real-time targets, and revised (up to current week) targets using a Recurrent Neural Network. Given a model $M$, for each observation week $t' \in \{1,2,\dots, t-1-k\}$, we concatenate the model's predictions $y(M,k)_{t'}$, real-time target $y_{t'+k}^{(t'+k)}$ as seen on observation week $t'$  and revised target $y_{t'+k}^{(t)}$ as $ C_{t'}^{t} = y(M,k)_{t'} \oplus y_{t'+k}^{(t'+k)} \oplus y_{t'+k}^{(t)}$,
where $\oplus$ is the concatenation operator.
A GRU is used to encode the sequence $\{C_{1}^{(t)}, \dots, C_{t-1-k}^{(t)}\}$:
\begin{equation}
    \{z_{1}^{(t)}, \dots, z_{t-1-k}^{(t)}\} = \mathrm{GRU}_{\mathrm{ME}}(\{C_{1}^{(t)}, \dots, C_{t-1-k}^{(t)}\})
\end{equation}

\par \noindent \textbf{\reiner.}
It leverages the information from above three modules of \model to refine model $M$'s prediction for current week $y(M,k)_{t}$. Specifically, it receives the latent encodings of signals $\{h_{i,t}^{(t+l)}\}_{i\in \fftall}$
from \bseqenc ,
$z_{t-k-1}^{(t)}$ from \modelenc, and the model's prediction $y(M,k)_t$ for week $t$.

\bseq encoding from different signals may have variable impact on refining the signal since a few signals may not very useful for current week's forecast (e.g., small revisions in mobility signals may not be important in some weeks). Moreover, because different models use signals from \dtst differently, we may need to focus on some signals over others to refine its prediction.
Therefore, we first take attention over \bseq encodings from all signals 
$\{h_{i,t}^{(t+l)}\}_{i\in \fftall}$ 
w.r.t $y(M,k)_t$. We use multiplicative attention mechanism with parameter $w \in \mathrm{R}^m$ based on \cite{Vaswani2017AttentionIA}:

\vspace{-5pt}
{\small
\begin{equation}
    \alpha_{i} = \mathrm{softmax}_i(y(M,k)_t w_h^T h_{i,t}^{t+l}),
    \quad\quad\quad 
    \bar{h}^{(t)} = \sum_{i \in \fftall}\left( \alpha_i h_{i,t}^{t+l} \right).
\end{equation}}
Finally we combine $\bar{h}^{(t)}$ and  $z_{t-k-1}^{(t)}$ through a two layer feed-forward layer $\mathrm{FNN}_{\mathrm{RF}}$ which outputs a 1-dim value followed by $\tanh$ activation to get the correction $\gamma_t \in [-1,1]$ i.e., $\gamma_t = \tanh(\mathrm{FFN}_{\mathrm{RF}}(\bar{h}^{(t)} \oplus z_{t-k-1}^{(t)}))$.
Finally, the refined prediction is $y^*(M,K)_t = (\gamma+1) y(M,K)_t$. Note that we limit the correction by \model by at most the magnitude of model's prediction because the average \be of targets is $4.9\%$ and less than $0.6\%$ of them have \be over 1. Therefore, we limit the refinement of prediction to this range. 

\par \noindent \textbf{Training:}
\label{sec:training}
There are two steps of training involved for \model: 1) model agnostic autoregressive \bseq prediction task to pre-train \bseqenc; 2) model-specific training for \probf.

\noindent \textit{Autoregressive \bseq prediction:} Pre-training on auxiliary tasks to improve the quality of latent embedding is a well-known technique for deep learning methods \citep{Devlin2019BERTPO,radford2018improving}.
We pre-train \bseqenc to predict the next values of backfill sequences $\{x_{t',i}^{(t'_r+1)}\}_{i\in \fftall}$. 
Note that we only use \bseq sequences $\{\bseq_{t}(t',i)\}_{i\in \fftall, t'<t}$ available till current week $t$ for training \bseqenc. The training procedure in itself is similar to Seq2Seq prediction problems~\citep{Sutskever2014SequenceTS} where for initial epochs we use the ground truth inputs at each step (teacher forcing) and then transition to using output predictions of previous time step by the recurrent module as input to next time step. Once we pre-train \bseqenc, we can use it for  \probf as well as \prob{} for current week $t$ \emph{for any model} $M$. Fine-tuning usually takes less than half the epochs required for pre-training enabling quick refinement of multiple models in parallel.

\noindent \textit{Model specific end-to-end training:} Given the pre-trained \bseqenc, we train, end-to-end, the parameters of all modules of \model. The training set consists of past model predictions $\langle y(M,k)_1, y(M,k)_2,\dots y(M,k)_{t-1} \rangle$ and backfill sequences $\{\bseq_{t}(t',i)\}_{i\in \fftall, t'\leq t}$. 
For datapoint $y(M,k)_{t'}$ of week $t'<t-k$, we input backfill sequences of signals whose observation week is $t'$ into \bseqenc to 
get latent encodings $\{h_{t',i}^{(t)}\}_{i\in \fftall}$. 
We also derive $z_{t'}^{(t)}$ from \modelenc and  finally \reiner ingests $z_{t'}^{(t)}$, $\{h_{t',i}^{t}\}_{i\in \fftall}$ and $y(M,k)_{t'}$ to get $\gamma_{t'}$.
Overall, we optimize the loss function:
 $   \mathcal{L}^{(t)} = \sum_{i=1}^{t-k-1}\left(\gamma_{t'} y(M,k)_{t'} - y_{t'+k}^{(t)} \right)^2
$.
 Following real-time forecasting, we train \model each week from scratch (including pre-training). Throughout training and forecasting for week $t$, we use $G_t$ as input to \bseqenc since it captures average similarities in \bseqs till current week $t$.
\section{Back2Future Experimental Results}
\label{sec:results}

In this section,  we describe a detailed empirical study to evaluate the effectiveness of our framework \model.
All experiments were run in an Intel i7 4.8 GHz CPU with Nvidia Tesla A4 GPU. The model typically takes around 1 hour to train for all regions. 
The appendix contains additional details (all hyperparameters and results for June-Dec 2020 and $k=1,3$ and GDP forecasting).


\noindent\textbf{Setup:} 
We perform real-time forecasting of COVID-19 related mortality (\fdt) for 50 US states. We leveraged observations (Section 2) from \bseq for period June 2020 - Dec. 2020 to design \model. We tuned the model hyperparameters using data from June 2020 to Aug. 2020  and tested it on the rest of dataset including completely \emph{unseen data} from Jan. 2021 to June 2021. For each week $t$, we train the model using the \dtst  dataset available till the current week $t$ (including \bseqs for all signals revised till $t$) for training. As described in Section \ref{sec:training}, for each week, we first pre-train \bseqenc on \bseq data and then train all components of \model for each model we aim to refine. Then, we predict the forecasts \fdt $y^*(M,k)_t$ for each model $M$. Similarly we also evaluated \model for real-time GDP forecasting task with detailed results in Appendix.
We observed that setting hyperparameter $\tau = c|\fftall|$ where $c\in \{2,3,4,5\}$ provided best results. Note that $\tau$ influences the sparsity of the graph as well as the efficiency of the model since sparser graphs lead to fast inference across $\mathrm{GConv}$ layers.
We also found setting $l=5$ provided the best performance.

\noindent\textbf{Evaluation:}
 We evaluate  the refined prediction $y^*(M,K)_t$ against the most revised version of the target, i.e. $y_{t'+k}^{(t_f)}$. In our case, $t_f$ is second week of Feb 2021.
For evaluation, we use standard metrics in this domain~\cite{reich_collaborative_2019,adhikari2019epideep}. Let absolute error of prediction $e(M,k)_t = |y_{t+k}^{(t_f)} - y^*(M,K)_t|$ for a week $t$ and model $M$.
We use (a) Mean Absolute Error $\mathrm{MAE}(M)=\frac{1}{T'} \sum_{i=1}^{T'} \mid e(M,k)_t \mid$ and (b) Mean Absolute Percentage Error $\mathrm{MAPE}=\frac{1}{T'} \sum_{i=1}^{T'}  {e(M,k)_t}~/~{\mid y_{t+k}^{(t_f)}\mid} $.

\vspace{-2pt}
\noindent\textbf{Candidate models:} We focus on refining/rectifying the top models from the \hub described in Section \ref{sec:obs}; these represent different variety of statistical and mechanistic models.

\begin{wraptable}[23]{l}{.75\linewidth}
\centering
\vspace{-0.2in}
\caption{\textit{\model consistently refines all models. \% improvements in MAE and MAPE scores averaged over all regions from Jan 2021 to June 2021}}
\label{tab:mainresult}
\vspace{-0.1in}
\scalebox{0.75}{
\begin{tabular}{c|l|r|r|r|r}
\multicolumn{1}{l|}{}                     &                         & \multicolumn{2}{c|}{\textbf{k=2}}                    & \multicolumn{2}{c|}{\textbf{k=4}}                   \\ \hline
\multicolumn{1}{l|}{\textbf{Cand. Model}} & \textbf{Refining Model} & \multicolumn{1}{l|}{MAE} & \multicolumn{1}{l|}{MAPE} & \multicolumn{1}{l|}{MAE} & \multicolumn{1}{l}{MAPE} \\ \hline
\multirow{5}{*}{\ensemble}                 & \ffnreg            & -0.35 $\pm$ 0.11            & -0.12 $\pm$ 0.22             & 0.87 $\pm$ 0.64             & 0.77 $\pm$ 0.14             \\ \cline{2-6} 
                                          & \modelreg            & -2.23 $\pm$ 0.82            & -1.57 $\pm$ 0.65             & -2.19 $\pm$ 0.35            & -2.85 $\pm$ 0.53            \\ \cline{2-6} 
                                          & \bseqregt           & -1.45 $\pm$ 0.14            & -2.73 $\pm$ 0.35             & -5.72 $\pm$ 0.21            & -6.72 $\pm$ 0.82            \\ \cline{2-6} 
                                          & \bseqreg          & 1.42 $\pm$ 0.60             & 0.37 $\pm$ 0.75              & 0.74 $\pm$ 0.36             & 0.44 $\pm$ 0.07             \\ \cline{2-6} 
                                          & Back2Future             & \textbf{5.25 $\pm$ 0.13}    & \textbf{4.39 $\pm$ 0.62}     & \textbf{4.41 $\pm$ 0.73}    & \textbf{3.15 $\pm$ 0.57}    \\ \hline
\multirow{5}{*}{\deepcovid}                    & \ffnreg            & -2.42 $\pm$ 0.22            & -1.51 $\pm$ 0.90             & -1.54 $\pm$ 0.57            & -0.48 $\pm$ 0.43            \\ \cline{2-6} 
                                          & \modelreg            & -3.02 $\pm$ 0.40            & -3.41 $\pm$ 0.16             & -2.91 $\pm$ 0.29            & -3.22 $\pm$ 0.74            \\ \cline{2-6} 
                                          & \bseqregt           & 2.24 $\pm$ 0.37             & 3.51 $\pm$ 0.21              & 1.93 $\pm$ 0.39             & 0.78 $\pm$ 0.37             \\ \cline{2-6} 
                                          & \bseqreg          & 2.13 $\pm$ 0.12             & 3.84 $\pm$ 0.78              & 1.08 $\pm$ 0.23             & 2.33 $\pm$ 0.97             \\ \cline{2-6} 
                                          & Back2Future             & \textbf{10.33 $\pm$ 0.19}   & \textbf{11.84 $\pm$ 0.18}    & \textbf{9.92 $\pm$ 0.98}    & \textbf{11.27 $\pm$ 0.88}   \\ \hline
\multirow{5}{*}{\yyg}                      & \ffnreg            & -2.08 $\pm$ 0.39            & -1.34 $\pm$ 0.12             & -2.64 $\pm$ 0.13            & -3.36 $\pm$ 0.18            \\ \cline{2-6} 
                                          & \modelreg            & -3.84 $\pm$ 0.08            & -6.99 $\pm$ 0.56             & -8.84 $\pm$ 0.96            & -5.61 $\pm$ 0.27            \\ \cline{2-6} 
                                          & \bseqregt           & -1.25 $\pm$ 0.70            & -0.7 $\pm$ 0.90              & -6.13 $\pm$ 0.08            & -5.31 $\pm$ 0.06            \\ \cline{2-6} 
                                          & \bseqreg          & -1.78 $\pm$ 0.74            & -2.26 $\pm$ 0.83             & -0.79 $\pm$ 0.21            & -0.62 $\pm$ 0.27            \\ \cline{2-6} 
                                          & Back2Future             & \textbf{8.93 $\pm$ 0.26}    & \textbf{6.32 $\pm$ 0.44}     & \textbf{7.32 $\pm$ 0.42}    & \textbf{5.73 $\pm$ 0.66}    \\ \hline
\multirow{5}{*}{\umass}                 & \ffnreg            & -3.25 $\pm$ 0.38            & -5.74 $\pm$ 0.75             & -1.01 $\pm$ 0.18            & -5.28 $\pm$ 0.07            \\ \cline{2-6} 
                                          & \modelreg            & -8.2 $\pm$ 0.61             & -7.54 $\pm$ 0.26             & -6.49 $\pm$ 0.29            & -7.56 $\pm$ 0.30            \\ \cline{2-6} 
                                          & \bseqregt           & -2.16 $\pm$ 0.22            & -1.88 $\pm$ 0.67             & -2.15 $\pm$ 0.24            & -2.87 $\pm$ 0.34            \\ \cline{2-6} 
                                          & \bseqreg          & 1.58 $\pm$ 0.49             & 0.86 $\pm$ 0.17              & 0.36 $\pm$ 0.06             & 0.96 $\pm$ 0.82             \\ \cline{2-6} 
                                          & Back2Future             & \textbf{5.43 $\pm$ 0.51}    & \textbf{4.66 $\pm$ 0.63}     & \textbf{3.32 $\pm$ 0.76}    & \textbf{3.11 $\pm$ 0.29}    \\ \hline
\multirow{5}{*}{\cmu}                   & \ffnreg            & -5.24 $\pm$ 0.57            & -4.93 $\pm$ 0.39             & -3.12 $\pm$ 0.71            & -0.65 $\pm$ 0.81            \\ \cline{2-6} 
                                          & \modelreg            & -8.17 $\pm$ 0.34            & -8.21 $\pm$ 0.24             & -3.72 $\pm$ 0.32            & -6.11 $\pm$ 0.84            \\ \cline{2-6} 
                                          & \bseqregt           & -0.67 $\pm$ 0.69            & -0.57 $\pm$ 0.32             & -0.46 $\pm$ 0.07            & -1.77 $\pm$ 0.79            \\ \cline{2-6} 
                                          & \bseqreg          & 1.46 $\pm$ 0.33             & 1.05 $\pm$ 0.16              & 2.38 $\pm$ 0.43             & 2.26 $\pm$ 0.02             \\ \cline{2-6} 
                                          & Back2Future             & \textbf{7.5 $\pm$ 0.60}     & \textbf{8.04 $\pm$ 0.58}     & \textbf{5.73 $\pm$ 0.19}    & \textbf{6.22 $\pm$ 0.58}   
\end{tabular}
}
\end{wraptable}

\noindent \textbf{Baselines:} Due to the novel problem, there are no standard baselines. 
    (a) \ffnreg: train an FFN for regression task that takes as inputs model's prediction and real-time target to predict the stable target. 
    (b) \modelreg: use the \modelenc architecture and append a linear layer that takes encodings from \modelenc and model's prediction and train it to refine the prediction.
    (c) \bseqreg: similarly, only use \bseqenc architecture and append a linear layer that takes encodings from \bseqenc and model's prediction to predict the stable target.
    (d) \bseqregt: similar to \bseqreg but remove the graph convolutional layers and retain only RNNs.
Note that \ffnreg and \modelreg do not use revision data. \bseqreg and \bseqregt don't use past predictions of the model.

\noindent \textbf{Refining real-time model-predictions:}
 We compare the mean percentage improvement (i.e., decrease) in MAE and MAPE scores of \model refined predictions of diverse set of top models w.r.t stable targets over 50 US states\footnote{Results described are statistically significance due to Wilcox signed rank test ($\alpha=0.05$) over 5 runs}.
 First, we observe that \model is the only method, compared to baselines, that improves
 scores for \emph{all} candidate models (Table \ref{tab:mainresult}) showcasing the necessity of incorporating backfill information (unlike \ffnreg and \modelreg) and model prediction history (unlike \bseqreg and \bseqregt). 

 We achieve an impressive avg. improvements of 6.93\%  and 6.79\% in MAE and MAPE respectively with improvements decreasing with increasing $k$.
 Candidate models refined by \model show improvement of over 10\% in over 25 states and over 15\% in 5 states (NJ, LA, GA, CT, MD). 
Due to \model refinement, the improved predictions of  \cmu and \deepcovid, which are ranked 3rd and 4th in \hub, outperform all the models in the hub (except for \ensemble) with impressive 7.17\% and 4.13\% improvements in MAE respectively. \umass, ranked 2nd, is improved by 11.24\%.

 \begin{wrapfigure}[10]{l}{.5\linewidth}
     \vspace{-0.8cm}
    \hspace{-0.3cm}
    \centering
    \begin{tabular}{cc}
    \centering
    \includegraphics[width=0.47\linewidth]{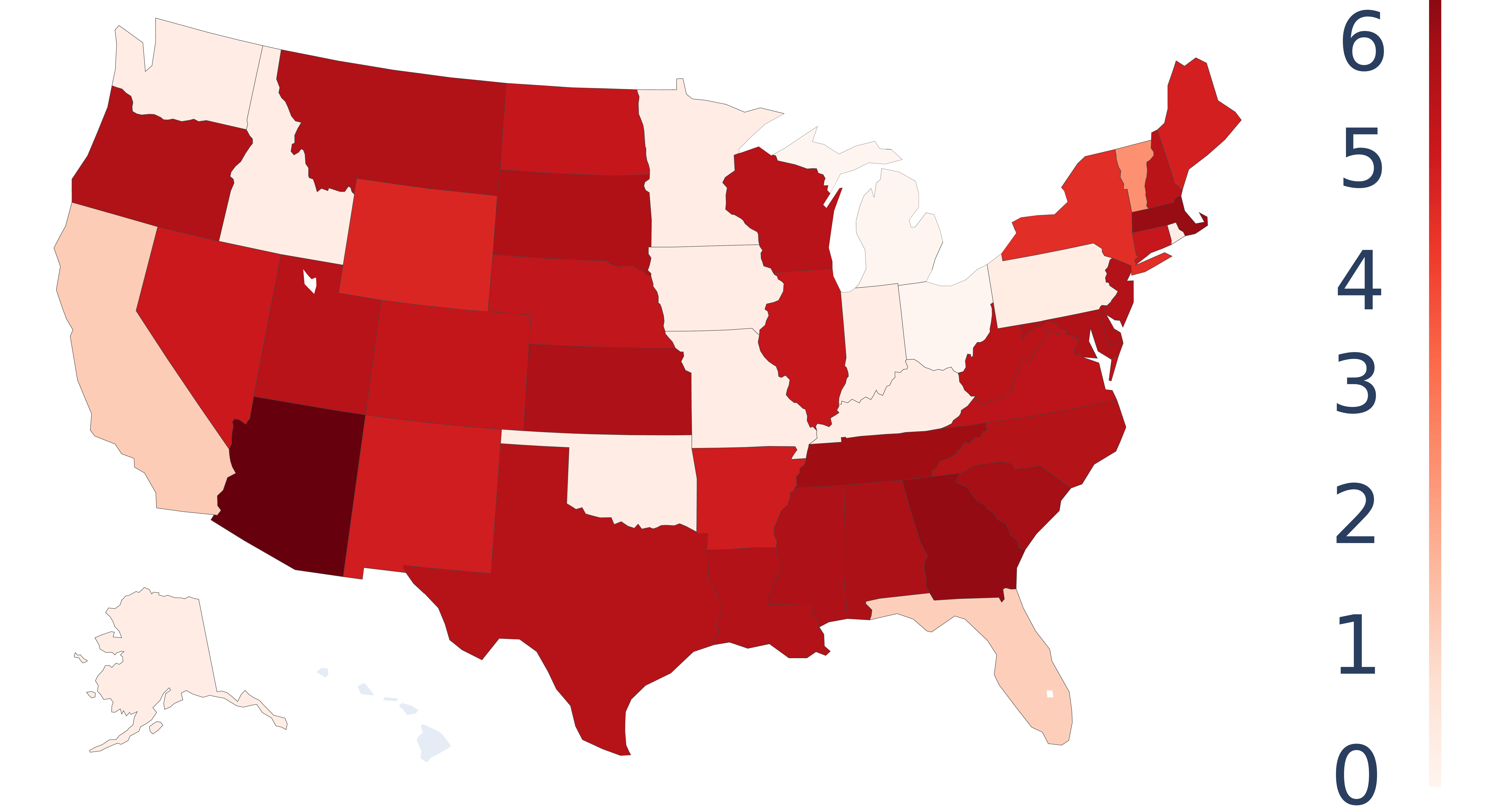} &
    \includegraphics[width=0.47\linewidth]{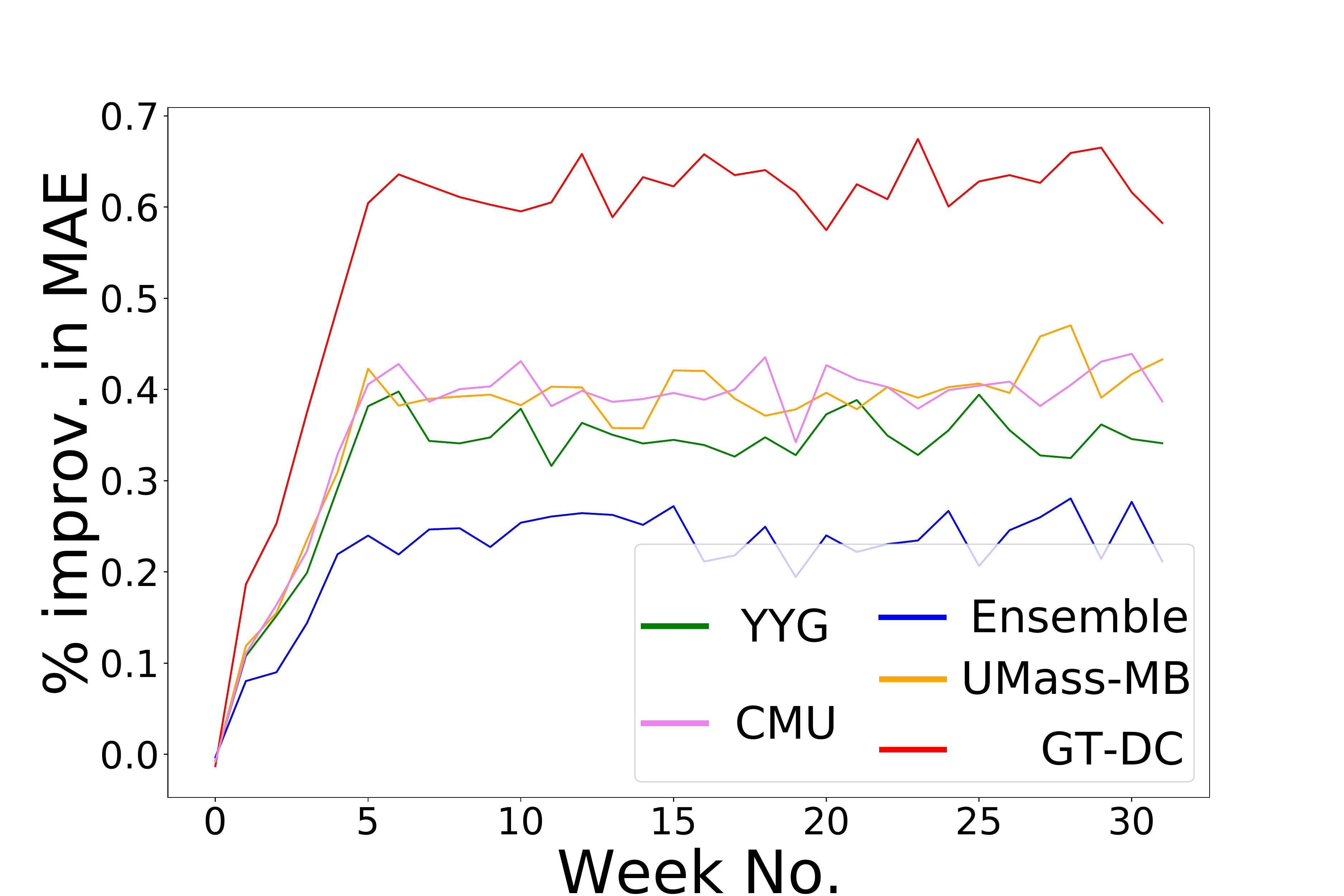} \\
    \makecell{(a) Average \% \\ decrease  of MAE} & 
    \makecell{ (b) \% improve. in\\ MAE for each week }
  \end{tabular}
  \vspace{-0.35cm}
    \caption{\textit{(a) \model refines \ensemble predictions significantly for most states. (b) efficacy of \model ramps up within 6 weeks of revision data.}}
    \label{fig:results}
\end{wrapfigure}
\model also improves \ensemble, which is the current best-performing 
model of the hub, by 3.6\% - 5.18\% 
with over 5\% improvement in 38 states, and with IL and TX experiencing over 15\% improvement.

\begin{wrapfigure}[9]{l}{.29\linewidth}
 \vspace{-.25in}
    \centering
    \includegraphics[width=.9\linewidth]{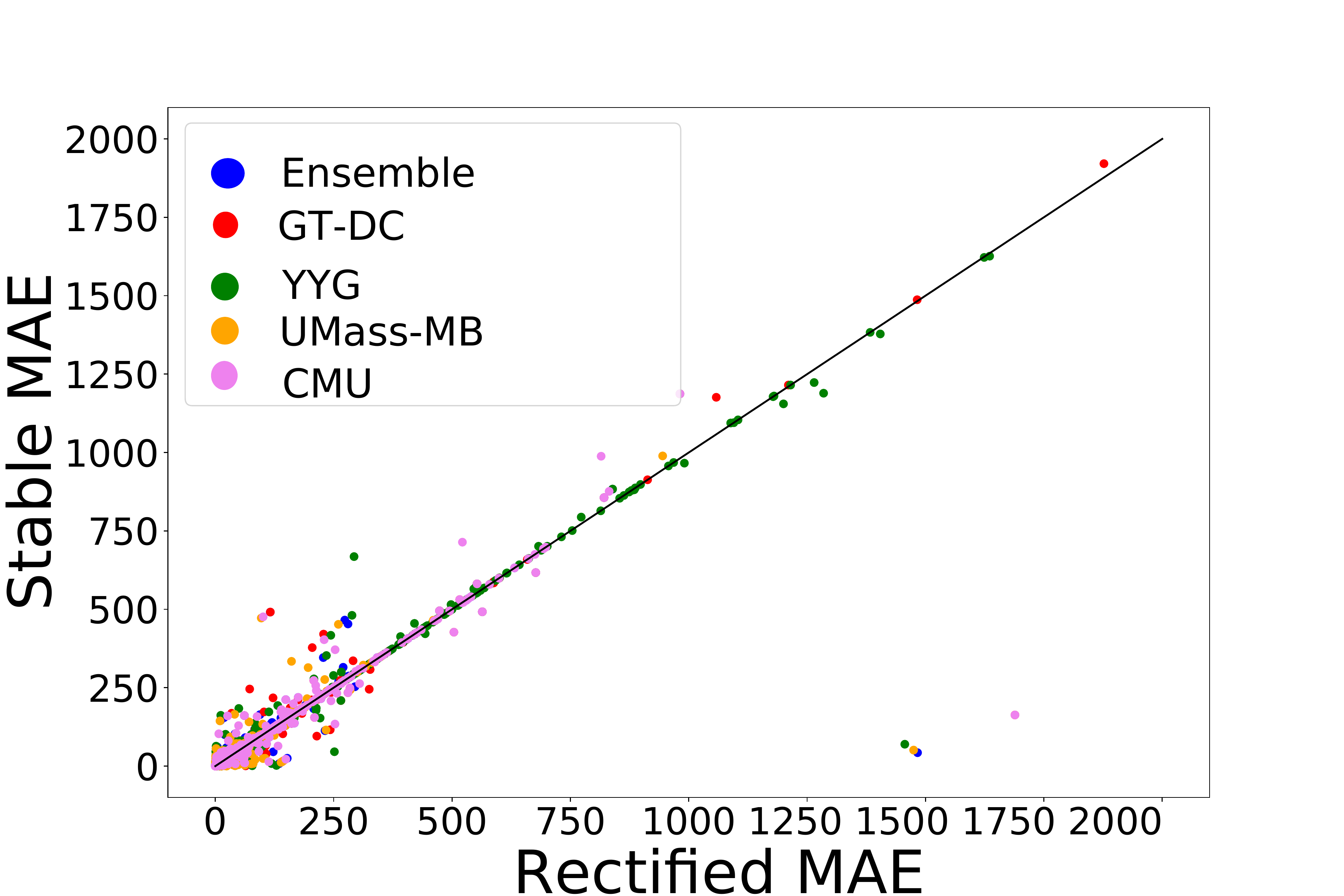}
    \caption{\textit{\model rectified MAE are closer to stable MAE}}
    \label{fig:rectify}
 \end{wrapfigure}

\noindent \textbf{Rectifying real-time model-evaluation:} We evaluate the efficacy of \model in rectifying the real-time evaluation scores for the \prob  problem. We noted in Obs \ref{obs:eval} that real-time MAE was lower than stable MAE by 9.6 on average.
The difference between \model rectified estimates MAE  stable MAE was reduced to 4.4, a 51.7\% decrease (Figure \ref{fig:rectify}). This results in increased MAE scores across most regions towards stable estimates. 
 Eg: We reduce the MAE difference in large highly populated states such as GA by 26.1\% (from 22.52 to 16.64) and TX by 90\% (from 10.8 to 1.04) causing an increase in MAE scores from real-time estimates by 5.88 and 9.4 respectively.

\noindent \textbf{Refinement as a function of data availability:} Note that during the initial weeks of the pandemic, we have access to very little revision data both in terms of length and number of \bseq. So we evaluate the mean performance improvement for each week across all regions (Figure \ref{fig:results}b).
 \model's performance ramps up and quickly stabilize in just 6 weeks. Since signals need around 4 weeks (Obs \ref{obs:bd}) to stabilize, this ramp-up time is impressive. Thus, \emph{\model needs a small amount of revision data to improve model performance.}

\par
\begin{wrapfigure}{r}{.6\linewidth}
    \vspace{-0.2in}
    \centering
    \begin{subfigure}{.38\linewidth}
    \centering
    \includegraphics[width=.93\linewidth]{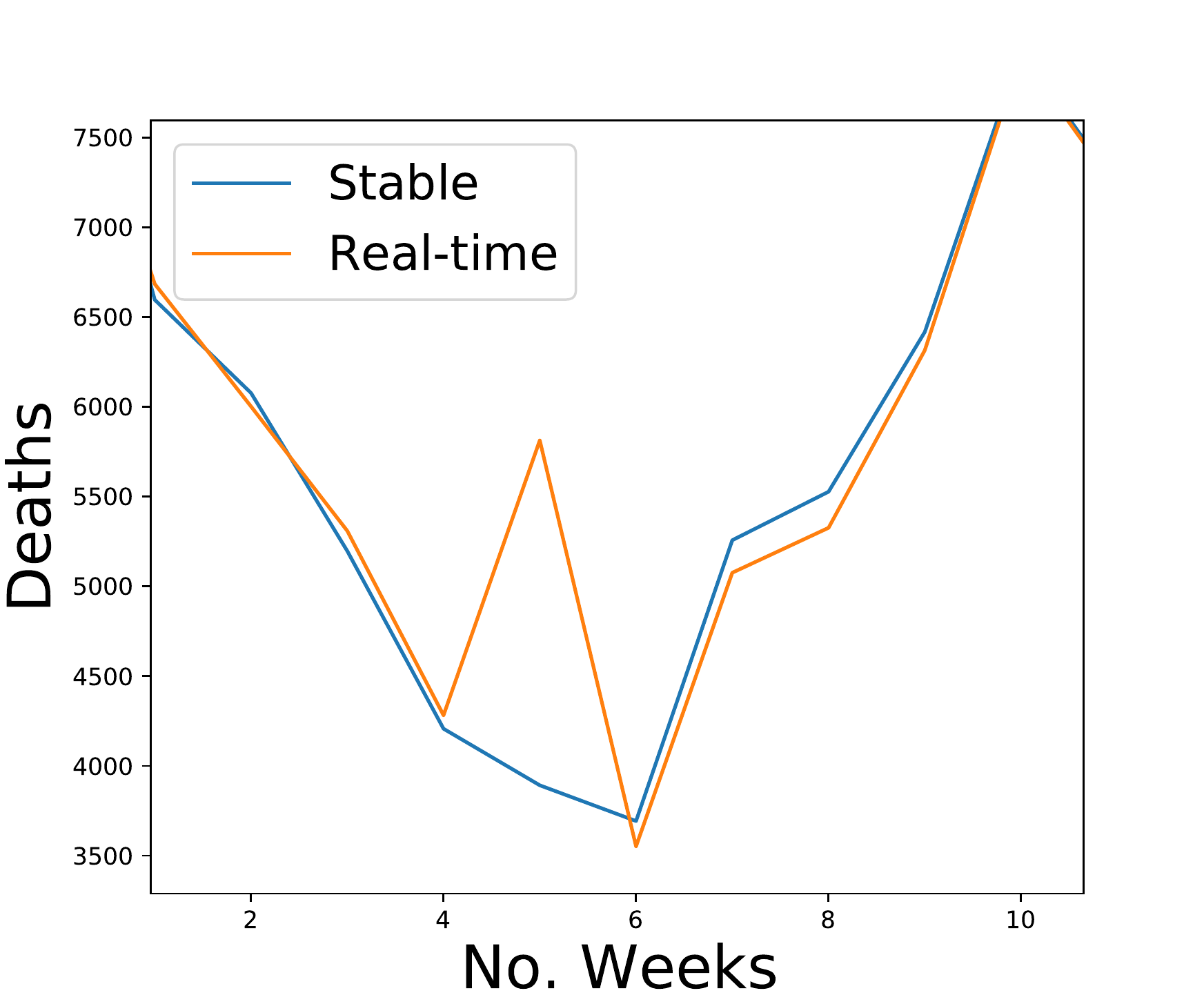}
    \caption{Revision of US deaths on week 5}
    \label{fig:usdeaths}
    \end{subfigure}\hfill
    \begin{subfigure}{.58\linewidth}
    \centering
    \includegraphics[width=.93\linewidth]{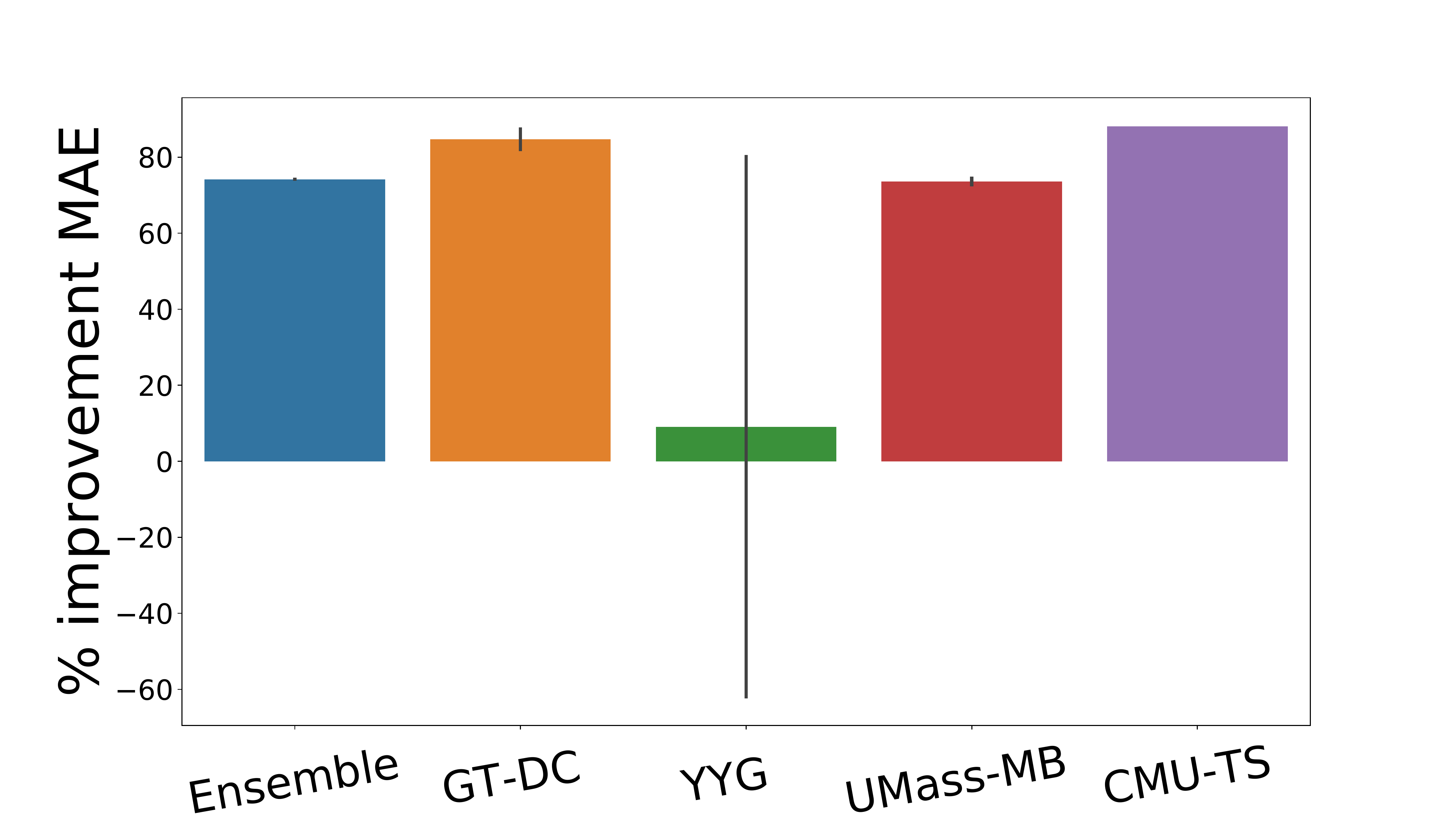}
    \caption{\% decrease in MAE due to \model refinement on week 5 targets}
    \label{fig:week5}
    \end{subfigure}
    \vspace{-5pt}
    \caption{\textit{\model adapts to abnormally high revisions}}
    \label{fig:anomaly}
    \vspace{-0.15in}
\end{wrapfigure}

 \noindent \textbf{\model adapts to anomalies}
During real-time forecasting, models need to be robust to anomalous data revisions. This is especially true during the initial stages of the pandemic when data collection is not fully streamlined and may experience large revisions.
Consider observation week 5 (first week of June) where there was an abnormally large revision to deaths nationwide (Figure \ref{fig:anomaly}a) when the \be was 48\%. \model still provided significant improvements for most model predictions (Figure \ref{fig:anomaly}b). Specifically, \ensemble's predictions are refined to be up to 74.2\% closer to stable target.

\noindent\textbf{\model refines GDP forecasts} To evaluate the extensibility of \model to other domains that encounter the problem of backfill, we tested on the task of forecasting US National GDP at 1 and 2 quarters into the future using 25 macroeconomic indicators from past and their revision history for years 2000-2021. We found that \model improves predictions of candidate models by 6\%-15\% and significantly outperforms baselines. The details of the dataset and results are found in Appendix Sections \ref{sec:gdp}, \ref{sec:gdp_res}.

\vspace{-10pt}
\section{Conclusion}
\vspace{-5pt}

We introduced and studied the challenging multi-variate backfill problem using COVID-19 and GDP forecasting as examples. We presented Back2Future (\model), the novel deep-learning method to model this phenomenon, which exploits our observations of cross-signal similarities using Graph Recurrent Neural Networks to refine predictions and rectify evaluations for a wide range of models. 
Our extensive experiments showed that leveraging similarity among backfill patterns via our proposed method leads to impressive 6 - 11 \% improvements in all the top models. 

As future work, our work can potentially help improve data collection and alleviate systematic differences in reporting capabilities across regions. 
We emphasize that we used the \dtst dataset as is, but our characterization can be helpful for anomaly detection~\citep{homayouni2021anomaly}. We can also study how backfill can affect uncertainty calibration in time-series analysis~\citep{pmlr-v119-yoon20c}. Adapting to situations where revision history is not temporally uniform across all features due to revisions of different frequency is another  research direction.

\section{Ethics Statement}

Although the features used in the \dtst dataset and for GDP forecasting are publicly available and anonymized without any sensitive information. However, due to the relevance of our dataset to public health and macroeconomics, prospects for misuse should not be discounted. The disparities in data collection across features and regions can also have implications on equity of prediction performance and is an interesting direction of research.
Our backfill refinement framework and \model is generalizable to any domain that deals with real-time prediction tasks with feature revisions. Therefore, there might be limited potential for misuse of our methods as well.
\section{Reproduciblity Statement}

As described in Section \ref{sec:results}, we evaluated our model over 5 runs with different random seeds to show the robustness of our results to randomization. We also provide a more extensive description of hyperparameters and data pre-processing in the Appendix.
The code for \model  and the \dtst  dataset is attached in the appendix. The dataset for GDP Forecasting is publicly available as described in Appendix Section \ref{sec:gdp}. Please refer to the README file in the code folder for more details to reproduce the results. On acceptance, we will also make the code and datasets available publicly to encourage reproducibility and allow for further exploration and research. 
\section{Acknowledgements}
This work was supported in part by the NSF (Expeditions CCF-1918770,
CAREER IIS-2028586, RAPID IIS-2027862, Medium IIS-1955883, Medium IIS-2106961,
CCF-2115126), CDC MInD program, ORNL, and faculty research awards from Facebook,
funds/computing resources from Georgia Tech.


\clearpage
\newpage
\bibliography{refs}
\bibliographystyle{iclr2022_conference}

\clearpage
\appendix
\textbf{\fontsize{15}{12}\selectfont Appendix for Back2Future: Leveraging Backfill Dynamics for Improving Real-time Predictions in Future}

Code for \model  and the \dtst  dataset is publicly available\footnote{\url{https://github.com/AdityaLab/Back2Future}}. Please refer to the README in the code folder for more details to reproduce the results.

\section{Additional Related work}
\noindent\textbf{Epidemic forecasting.} 
Broadly   two classes of approaches  have been devised: traditional mechanistic epidemiological models~\cite{shaman2012forecasting,zhang2017forecasting}, and the fairly newer statistical approaches~\cite{brooks_nonmechanistic_2018,adhikari2019epideep,osthus2019dynamic}, which have become among the top performing ones for multiple forecasting tasks~\cite{reich_collaborative_2019}. 

Statistical models have been helpful in using digital indicators such as
search queries~\cite{ginsberg2009detecting,yang2015accurate} and social media~\cite{culotta2010towards,lampos2010flu}, that can give more lead time than traditional surveillance methods.
Recently, deep learning models have seen much work. They can use heterogeneous and multimodal data and extract richer representations, including for modeling spatio-temporal dynamics~\cite{adhikari2019epideep,deng2020cola} and transfer learning~\cite{panagopoulos2020transfer,rodriguez_steering_2021}. Our work can be thought of refining any model (mechanistic/statistical) in this space. 

\noindent\textbf{Revisions and backfill.}
The topic of revisions has not received as much attention, with few exceptions. 
In epidemic forecasting, a few papers have mentioned about the `backfill problem' and its effects on performance~\citep{chakraborty_what_2018,rodriguez_deepcovid_2021,Altieri2021Curating,rangarajan2019forecasting} and evaluation~\citep{reich_collaborative_2019}; Some works proposed to address the problem via simple models like linear regression~\citep{chakraborty2014forecasting} or 'backcasting'~\citep{brooks_nonmechanistic_2018} the observed targets.
However they focus only on revisions in the \emph{target}, and study only in context of influenza forecasting, which is substantially less noisy and more regular than forecasting for the novel COVID-19 pandemic.   
\cite{reich_collaborative_2019} proposed a framework to study how backfill affects the evaluation of multiple models, but it is limited to label backfill and flu forecasting.
Other works use data assimilation and sensor fusion by leveraging revision free digital signals to refine noisy features for nowcasting \citep{hawryluk2021gaussian,farrow2016modeling,lampos2015advances, nunes2013nowcasting,osthus2019even}. However, we observed significant backfill in digital signals as well for COVID-19. Moreover, our model doesn't require revision-free data sources.
Some works model revision of event-based features as count variables which can't be applicable to many important features like mobility, exposure \citep{stoner2020multivariate,lawless1994adjustments}.
\cite{clements2019data} surveys several domain-specific~\citep{carriero2015forecasting} or essentially linear techniques in economics for data revision/correction behavior of the source of several macroeconomic indicators~\citep{croushore2011forecasting}.
In contrast, we study the more challenging problem of multi-variate backfill for both features and targets and  show how to leverage our insights for more general neural framework to \emph{improve} both model predictions and evaluation. 

\section{Details about GDP prediction task}
\label{sec:gdp}
\subsection{Dataset}

We curated the dataset collected from the Real-Time Data Research Center, Federal Reserve Bank Philadelphia which can be accessed publicly \footnote{Link to dataset: \url{https://www.philadelphiafed.org/surveys-and-data/real-time-data-research/real-time-data-set-full-time-series-history}}. We extracted the following important macroeconomic features denoted by tickers in parenthesis: Real GDP (ROUTPUT), Real Personal Consumption Expenditures Goods and Services(RCON, RCONG, RCONS), Real Gross Private Domestic Investment (RINVBF, RINVRESID), Real net Exports (RNX), Total Government Consumption (RG, RGF, RGSL), Household Consumption Expenditure (RCONHH), Final Consumption Expenditure (NCON), Nominal GDP (NOUTPUT), Nominal Personal Consumption Expenditures (NCON), Wage and Salary Disbursement (WSD), Other labor Income (OLI), Rental Income (RENTI), Dividends (DIV), Personal Income (PINTI), Transfer Payments (TRANR), Personal Saving Rate (RATESAV), Disposable Personal Income (NDPI), Unemployment Rate (RUC), Civilian Labor Force (LFC), Consumer Price Index (CPI).
\subsection{Setup}

For each of these 25 features, quarterly data is available since 1965 to third quarter of 2021 and the data is revised quarterly towards more accurate values. Our goal is to predict the Real GDP (ROUTPUT) $k$ quarters ahead using real-time data (data available till current quarter including past data revised till current quarter) where $k\in \{1, 2\}$. We tune the hyperparameters for model for using data from years 1995-2000 and test on the unseen period of 2000-2021. Due to lack of standard baselines, we chose as candidate models 1. Vector Autoregression (VAR) model, a standard model in macroeconmics literature \citep{robertson1999vector,runstler2003short,baffigi2004bridge}, 2. 4-layer feed forward network (FFN) on current quarter's features similar to \citep{tkacz1999forecasting} and a Recurrent Neural Network (GRU) on past GDP values.

\section{Nature of backfill dynamics (More details)}

\subsection{Considering Delay in reporting} Let current week be $t$.
Due to delay in reporting, the first observed value for some signals could be delayed by 1 to 3 weeks. For instance, $\bseq(i,t)$ may not have the first $\delta_i$ values due to $\delta_i$ weeks delay in reporting. In such cases, for analysis of \bseq in Section \ref{sec:obs}, we take $d_{i,t}^{t+\delta_i}$ as the first value of $\bseq(i,t) = \langle d_{i,t}^{t+\delta_i}, d_{i,t}^{t+\delta_i+1}, \cdots, d_{i,t}^{t_f} \rangle$. Subsequently, $\be(r,i,t)$ is only defined for $r>\delta_i$ as $\be(r,i,t) = \frac{|d_{i,t}^{(t+r)} - d_{i,t}^{t_f}|}{|d_{i,t}^{t_f}|}$.

\paragraph{Real-time forecasting} As mentioned in the main paper, we handle delays in reporting real-time data by approximating from the previous week.
During real-time forecasting, we cannot wait for $\delta_i$ weeks to get the first value of a signal. Therefore, we replace $d_{i,t}^{(t)}$ with the most revised value of last week for which the signal is available. Let $t'<t$ is the last week before $t$ for which we have $d_{i,t'}^{(t)}$. Then, we use $d_{i,t'}^{(t)}$ in place of unavailable $d_{i,t}^{(t)}$.  Note that such cases of missing initial value during real-time forecasting is very uncommon.

\subsection{Description of canonical backfill behaviours}
We saw in Observation \ref{obs:patterns} that clustering \bseqs resulted in 5 canonical behaviours.
The behaviors of these clusters as shown in Figure \ref{fig:patterns} can be described as:
\begin{enumerate*}
    \item \bed: \bseq values stabilizes quickly (within a week) to a lower stable value.
    \item \beinc: \bseq values increase in 2 to 5 weeks and stabilize.
    \item \bst: \bseq values either remain constant (no significant revision) or due to reporting errors there may be anomalous values in between. (E.g., for a particular week in the middle of \bseq the signals are revised to 0 due to reporting error).
    \item \blin: \bseq values increase and stabilize very late during revision.
    \item \bmd: \bseq values change and stabilize after 8 to 10 weeks of revision.
\end{enumerate*}

\begin{table}[h]
\centering
\caption{Pearson Correlation Coefficient (PCC) between \be and \revmae using stable labels over real-time labels}
\begin{tabular}{|l|l|}
\hline
\textbf{Model}    & \textbf{PCC} \\ \hline
\ensemble & -0.327       \\ \hline
\deepcovid      & -0.322       \\ \hline
\yyg   & 0.110        \\ \hline
\umass   & -0.291       \\ \hline
\cmu    & -0.468       \\ \hline
\end{tabular}
\label{tab:pcc}
\vspace{-0.26in}
\end{table}
\subsection{Correlation between \be and model performance}
In Observation \ref{obs:model},
we found that models  differ in how their performance is impacted by \be on the label and found that relationship between \be and difference in MAE (\revmae) was varied across models with some models (\yyg) even showing positive correlation between \be and \revmae.
To further study this relationship between \be and reduction in MAE on using \emph{stable} labels for evaluation over \emph{real-time} labels by computing the Pearson correlation coefficient (PCC) between \be and \revmae for in Table \ref{tab:pcc}. As seen from Figure \ref{fig:obs_corr}, we observe that for \yyg, PCC is positive, indicating that \yyg's scores are actually due to larger \be on labels. We also see significant differences in PCC for \yyg and \cmu in over other models.

\section{Hyperparameters}
In this section, we describe in detail hyperparameters related to data preprocessing and \model architecture.
\subsection{Data Pre-processing}
\paragraph{Missing real-time data} Sometimes there is a delay in receiving signals for the current week. In this case, we use values from the most revised version of last observation week for real-time forecasting.
\paragraph{Missing revisions} Once we start observing a signal $d_{t}^{(t)}$ from week $t$ till $t_f$, there are weeks $t'$ in between where the revised value of this signals is not available or value received is zero. This gives rise to \emph{Spike} behaviour (Figure \ref{fig:patterns}). Before training, we replace this value with the previous value of \bseq.
\paragraph{Termination of revisions} We observed that for some digital signals, revisions stop a few months in the future. This could be due to the termination of data correction for that signal for older observation weeks. In such cases, we assume that signals are stabilized and use the last available revised values to fill the following values of \bseq.
\paragraph{Scaling signal values} Since each signal that is received has a very different range of values, we rescale each signal with a mean 0. and standard deviation 1.0. Note that since \model is trained separately for each week, we normalize the data for each week before training.
\subsection{Architecture}

\paragraph{\bseqenc} The dimension size of all latent encodings $h_{i,t'}^{(t'_r)}$ and $v_{i,t'}^{(t'_r)}$ is set to 50. Each $GConv$ is a single layer of graph convolutional neural network with weight matrix of size $\mathrm{R}^{50\times 50}$.

\paragraph{\modelenc} For $GRU_{ME}$ We use a GRU of a single hidden layer with output size 50.

\paragraph{\reiner} the feed-forward network $FFN_{RF}$is a 2 layer network with hidden layers of size 60 and 30 followed by final layer outputting 1-dimensional scalar.

\paragraph{Training hyperparameters} We used a learning rate of $10^{-3}$ for pre-training and $5\times 10^{-4}$ for fine-tuning.
The pretraining task usually around takes 2000 epoch to train with the first 1000 epochs using teacher forcing and each of the rest of the epoch using teacher forcing with a probability of 0.5. Fine-tuning takes between 500 to 1000 epoch depending on the model, region, and week of the forecast.

As mentioned in Section 4, we used data from June 2020 to Dec 2020 for model design using Observations from Section 2. The hyperparameter tuning was done using data from June 2020 to Aug 2020 and evaluated for time period of Jan 2021 to July 2021. 
Overall, we found that most hyperparameters are not sensitive. The most sensitive ones mentioned in the main paper are $c\in \{2,3,4,5\}$ that controls sparsity of graph $G_t$ and $l=5$ that controls how many steps we auto-regress using \bseqenc to derive latent encodings for \bseq during inference.

\section{Additional Results}
\subsection{COVID-19 Forecasting}
We show the average \% improvements of all baselines and \model in Table \ref{tab:ap_preds} including for $k=1,3$ week ahead forecasts (We show for $k=2,4$ in main paper Table 2 as well).
We show the results for the both June 2020 to Dec 2020 data, which was observed during model design and Jan 2021 to June 2021 which was unseen. \model show similar performance for both time periods.
\model clearly outperforms all baselines and provides similar improvements for \hub models for $k=1$ week ahead forecasts as described in Section \ref{sec:results} for other values of $k$.
\begin{table}[h]
\caption{\% improvement in MAE and MAPE scores averaged over all regions from June 2020 to Dec 2020}
\label{tab:ap_preds}
\scalebox{0.85}{
\begin{tabular}{|l|l|l|l|l|l|l|l|l|l|}
\hline
                                   &                & \multicolumn{2}{c|}{k=1}        & \multicolumn{2}{c|}{k=2}        & \multicolumn{2}{c|}{k=3}        & \multicolumn{2}{c|}{k=4}      \\ \hline
Cand. Model                        & Refining Model & \textbf{MAE}   & \textbf{MAPE}  & \textbf{MAE}   & \textbf{MAPE}  & \textbf{MAE}   & \textbf{MAPE}  & \textbf{MAE}  & \textbf{MAPE} \\ \hline
\multirow{5}{*}{\textbf{\ensemble}} & \ffnreg   & -0.21          & -0.45          & -0.35          & -0.12          & 0.48           & -0.29          & 0.81          & 0.36          \\ \cline{2-10} 
                                   & \modelreg   & -1.67          & -0.61          & -2.23          & -1.57          & -1.51          & -3.78          & -1.13         & -2.36         \\ \cline{2-10} 
                                   & \bseqregt  & 0.15           & 0.19           & -1.45          & -2.73          & -1.74          & -3.21          & -5.2          & -4.78         \\ \cline{2-10} 
                                   & \bseqreg & 2.1            & 0.29           & 1.42           & 0.37           & 0.37           & 0.22           & 0.72          & 0.28          \\ \cline{2-10} 
                                   & \model    & \textbf{6.22}  & \textbf{6.13}  & \textbf{5.18}  & \textbf{5.47}  & \textbf{3.64}  & \textbf{3.9}   & \textbf{3.61} & \textbf{6.3}  \\ \hline
\multirow{5}{*}{\textbf{\deepcovid}}    & \ffnreg   & -1.92          & -1.45          & -2.42          & -1.51          & -2.04          & -6.07          & -1.86         & -0.49         \\ \cline{2-10} 
                                   & \modelreg   & -3.9           & -4.44          & -3.02          & -3.41          & -2.62          & -3.95          & -2.81         & -3.26         \\ \cline{2-10} 
                                   & \bseqregt  & 0.51           & 0.46           & 2.24           & 3.51           & 2.31           & 2.42           & 1.6           & 0.57          \\ \cline{2-10} 
                                   & \bseqreg & 2.92           & 2.5            & 2.13           & 3.84           & 3.6            & 3.55           & 1.42          & 2.27          \\ \cline{2-10} 
                                   & \model    & \textbf{10.44} & \textbf{12.37} & \textbf{12.68} & \textbf{12.59} & \textbf{11.42} & \textbf{10.67} & \textbf{9.64} & \textbf{8.97} \\ \hline
\multirow{5}{*}{\textbf{YYG}}      & \ffnreg   & -3.45          & -1.53          & -2.08          & -1.34          & -4.62          & -3.89          & -1.37         & -3.06         \\ \cline{2-10} 
                                   & \modelreg   & -1.47          & -0.92          & -3.84          & -6.99          & -5.57          & -4.16          & -9.29         & -5.81         \\ \cline{2-10} 
                                   & \bseqregt  & -1.39          & -0.59          & -1.25          & -0.7           & -2.57          & -3.72          & -6.65         & -5.45         \\ \cline{2-10} 
                                   & \bseqreg & -1.98          & -1.92          & -1.78          & -2.26          & -1.99          & -1.53          & -0.61         & -0.43         \\ \cline{2-10} 
                                   & \model    & \textbf{10.64} & \textbf{7.74}  & \textbf{8.84}  & \textbf{5.98}  & \textbf{7.04}  & \textbf{7.64}  & \textbf{6.8}  & \textbf{5.27} \\ \hline
\multirow{5}{*}{\textbf{\umass}} & \ffnreg   & -2.37          & -2.03          & -3.25          & -5.74          & -2.16          & -3.17          & -1.44         & -4.84         \\ \cline{2-10} 
                                   & \modelreg   & -3.52          & -4.61          & -8.2           & -7.54          & -5.19          & -7.82          & -5.99         & -7.43         \\ \cline{2-10} 
                                   & \bseqregt  & -1.01          & -0.92          & -2.16          & -1.88          & -2.13          & -0.67          & -2.29         & -2.31         \\ \cline{2-10} 
                                   & \bseqreg & 0.92           & 0.78           & 1.58           & 0.86           & -1.26          & 0.45           & 0.06          & 0.03          \\ \cline{2-10} 
                                   & \model    & \textbf{4.44}  & \textbf{5.31}  & \textbf{5.21}  & \textbf{4.92}  & \textbf{3.25}  & \textbf{4.74}  & \textbf{3.94} & \textbf{3.49} \\ \hline
\multirow{5}{*}{\textbf{\cmu}}   & \ffnreg   & -2.22          & -4.17          & -5.24          & -4.93          & -2.87          & -1.95          & -3.19         & -6.7          \\ \cline{2-10} 
                                   & \modelreg   & -6.59          & -4.11          & -8.17          & -8.21          & -3.32          & -6.3           & -3.75         & -9.1          \\ \cline{2-10} 
                                   & \bseqregt  & 0.46           & 0.71           & -0.67          & -0.57          & -0.73          & -0.19          & -0.38         & -0.12         \\ \cline{2-10} 
                                   & \bseqreg & 1.76           & 2.34           & 1.46           & 1.05           & 2.74           & 2.47           & 2.11          & 2.84          \\ \cline{2-10} 
                                   & \model    & \textbf{7.54}  & \textbf{8.22}  & \textbf{8.75}  & \textbf{10.48} & \textbf{5.84}  & \textbf{7.62}  & \textbf{6.93} & \textbf{6.28} \\ \hline
\end{tabular}
}

\end{table}

\begin{table}[h]
\centering
\caption{\% improvement in MAE and MAPE scores averaged over all regions from Jan 2021 to June 2021}
\label{tab:ap_preds2}
\scalebox{0.85}{
\begin{tabular}{|c|l|r|r|r|r|r|r|r|r|}
\hline
\multicolumn{1}{|l|}{}             &                & \multicolumn{2}{c|}{k=1}                                               & \multicolumn{2}{c|}{k=2}                                               & \multicolumn{2}{c|}{k=3}                                               & \multicolumn{2}{c|}{k=4}                                               \\ \hline
\multicolumn{1}{|l|}{Cand. Model}  & Refining Model & \multicolumn{1}{l|}{\textbf{MAE}} & \multicolumn{1}{l|}{\textbf{MAPE}} & \multicolumn{1}{l|}{\textbf{MAE}} & \multicolumn{1}{l|}{\textbf{MAPE}} & \multicolumn{1}{l|}{\textbf{MAE}} & \multicolumn{1}{l|}{\textbf{MAPE}} & \multicolumn{1}{l|}{\textbf{MAE}} & \multicolumn{1}{l|}{\textbf{MAPE}} \\ \hline
\multirow{5}{*}{\textbf{\ensemble}} & \ffnreg   & -0.26                             & -0.55                              & -0.35                             & -0.12                              & 0.43                              & -0.29                              & 0.87                              & 0.77                               \\ \cline{2-10} 
                                   & \modelreg   & -1.75                             & -0.74                              & -2.23                             & -1.57                              & -1.63                             & -3.66                              & -2.19                             & -2.85                              \\ \cline{2-10} 
                                   & \bseqregt  & 0.08                              & 0.27                               & -1.45                             & -2.73                              & -1.42                             & -3.88                              & -5.72                             & -6.72                              \\ \cline{2-10} 
                                   & \bseqreg & 1.85                              & 0.28                               & 1.42                              & 0.37                               & 0.65                              & 0.28                               & 0.74                              & 0.44                               \\ \cline{2-10} 
                                   & \model    & \textbf{6.31}                     & \textbf{6.04}                      & \textbf{5.25}                     & \textbf{4.39}                      & \textbf{3.31}                     & \textbf{4.19}                      & \textbf{4.41}                     & \textbf{3.15}                      \\ \hline
\multirow{5}{*}{\textbf{\deepcovid}}    & \ffnreg   & -2.66                             & -1.35                              & -2.42                             & -1.51                              & -1.49                             & -6.65                              & -1.54                             & -0.48                              \\ \cline{2-10} 
                                   & \modelreg   & -3.51                             & -4.21                              & -3.02                             & -3.41                              & -2.72                             & -4.61                              & -2.91                             & -3.22                              \\ \cline{2-10} 
                                   & \bseqregt  & 0.26                              & 0.67                               & 2.24                              & 3.51                               & 2.34                              & 2.54                               & 1.93                              & 0.78                               \\ \cline{2-10} 
                                   & \bseqreg & 2.29                              & 2.95                               & 2.13                              & 3.84                               & 2.37                              & 3.52                               & 1.08                              & 2.33                               \\ \cline{2-10} 
                                   & \model    & \textbf{9.61}                     & \textbf{10.2}                      & \textbf{10.33}                    & \textbf{11.84}                     & \textbf{10.79}                    & \textbf{10.93}                     & \textbf{9.92}                     & \textbf{11.27}                     \\ \hline
\multirow{5}{*}{\textbf{YYG}}      & \ffnreg   & -3.62                             & -1.53                              & -2.08                             & -1.34                              & -4.41                             & -3.88                              & -2.64                             & -3.36                              \\ \cline{2-10} 
                                   & \modelreg   & -1.35                             & -1.69                              & -3.84                             & -6.99                              & -5.31                             & -4.6                               & -8.84                             & -5.61                              \\ \cline{2-10} 
                                   & \bseqregt  & -1.74                             & -0.56                              & -1.25                             & -0.7                               & -2.85                             & -2.31                              & -6.13                             & -5.31                              \\ \cline{2-10} 
                                   & \bseqreg & -2.62                             & -1.84                              & -1.78                             & -2.26                              & -2.67                             & -1.17                              & -0.79                             & -0.62                              \\ \cline{2-10} 
                                   & \model    & \textbf{9.52}                     & \textbf{6.14}                      & \textbf{8.93}                     & \textbf{6.32}                      & \textbf{6.82}                     & \textbf{7.58}                      & \textbf{7.32}                     & \textbf{5.73}                      \\ \hline
\multirow{5}{*}{\textbf{\umass}} & \ffnreg   & -2.56                             & -1.44                              & -3.25                             & -5.74                              & -1.46                             & -2.58                              & -1.01                             & -5.28                              \\ \cline{2-10} 
                                   & \modelreg   & -3.88                             & -3.85                              & -8.2                              & -7.54                              & -5.55                             & -8.12                              & -6.49                             & -7.56                              \\ \cline{2-10} 
                                   & \bseqregt  & -1.83                             & -0.91                              & -2.16                             & -1.88                              & -2.73                             & -0.76                              & -2.15                             & -2.87                              \\ \cline{2-10} 
                                   & \bseqreg & 0.66                              & 0.73                               & 1.58                              & 0.86                               & -1.12                             & 0.36                               & 0.36                              & 0.96                               \\ \cline{2-10} 
                                   & \model    & \textbf{4.51}                     & \textbf{4.75}                      & \textbf{5.43}                     & \textbf{4.66}                      & \textbf{5.68}                     & \textbf{4.89}                      & \textbf{3.32}                     & \textbf{3.11}                      \\ \hline
\multirow{5}{*}{\textbf{\cmu}}   & \ffnreg   & -3.11                             & -4.46                              & -5.24                             & -4.93                              & -3.52                             & -2.32                              & -3.12                             & -0.65                              \\ \cline{2-10} 
                                   & \modelreg   & -6.81                             & -5.92                              & -8.17                             & -8.21                              & -3.75                             & -7.6                               & -3.72                             & -6.11                              \\ \cline{2-10} 
                                   & \bseqregt  & 0.41                              & 0.85                               & -0.67                             & -0.57                              & -0.73                             & -0.29                              & -0.46                             & -1.77                              \\ \cline{2-10} 
                                   & \bseqreg & 1.33                              & 2.46                               & 1.46                              & 1.05                               & 2.17                              & 2.17                               & 2.38                              & 2.26                               \\ \cline{2-10} 
                                   & \model    & \textbf{7.54}                     & \textbf{8.22}                      & \textbf{7.5}                      & \textbf{8.04}                      & \textbf{6.42}                     & \textbf{8.23}                      & \textbf{5.73}                     & \textbf{6.22}                      \\ \hline
\end{tabular}
}
\end{table}

\subsection{Real-time GDP Forecasting}
\label{sec:gdp_res}
We also compare performance of \model with the baselines for GDP forecasting task in Table \ref{tab:gdp_all}.
\begin{table}[h]
    \centering
    \caption{\% improvement in MAE and MAPE scores for GDP Forecasting from 2000 to 2021}
    \label{tab:gdp_all}
    \scalebox{0.85}{

\begin{tabular}{|c|l|l|l|l|l|}
\hline
\multicolumn{1}{|l|}{}            &                & \multicolumn{2}{c|}{\textbf{k=1}}             & \multicolumn{2}{c|}{\textbf{k=2}}           \\ \hline
\multicolumn{1}{|l|}{Cand. Model} & Refining Model & \textbf{MAE}           & \textbf{MAPE}        & \textbf{MAE}         & \textbf{MAPE}        \\ \hline
\multirow{5}{*}{VAR}              & \ffnreg   & 2.44$\pm$0.66             & 2.22 $\pm$ 0.40         & 3.09$\pm$0.39           & 4.23$\pm$0.24           \\ \cline{2-6} 
                                  & \modelreg   & 4.05$\pm$ 0.32            & 3.73$\pm$0.13           & 3.48$\pm$ 0.23          & 3.04$\pm$0.29           \\ \cline{2-6} 
                                  & \bseqregt  & 3.32$\pm$0.23             & 2.40$\pm$0.81           & 3.39$\pm$0.14           & 2.99$\pm$0.48           \\ \cline{2-6} 
                                  & \bseqreg & 6.36$\pm$0.38             & 6.11$\pm$0.15           & 5.54$\pm$0.42           & 4.95$\pm$0.23           \\ \cline{2-6} 
                                  & \model    & \textbf{15.88 $\pm$ 0.18} & \textbf{15.71$\pm$0.63} & \textbf{14.94$\pm$0.34} & \textbf{10.26$\pm$0.56} \\ \hline
\multirow{5}{*}{FFN}              & \ffnreg   & -0.11$\pm$0.48            & -0.15$\pm$0.74          & 0.48$\pm$0.83           & 0.36$\pm$0.50           \\ \cline{2-6} 
                                  & \modelreg   & 0.28$\pm$0.44             & -0.07$\pm$0.13          & -0.24$\pm$0.48          & -0.23$\pm$0.62          \\ \cline{2-6} 
                                  & \bseqregt  & 2.26$\pm$0.43             & 1.98$\pm$0.85           & 2.29$\pm$0.18           & 2.20$\pm$0.36           \\ \cline{2-6} 
                                  & \bseqreg & 2.85$\pm$0.51             & 2.55$\pm$0.39           & 2.39$\pm$0.13           & 3.77$\pm$0.42           \\ \cline{2-6} 
                                  & \model    & \textbf{6.63 $\pm$ 0.13}  & \textbf{7.30$\pm$0.33}  & \textbf{6.07$\pm$0.44}  & \textbf{6.61$\pm$ 0.52} \\ \hline
\multirow{5}{*}{GRU}              & \ffnreg   & -1.13 $\pm$ 0.03          & 0.82$\pm$0.26           & -1.92$\pm$0.42          & -0.70$\pm$0.21          \\ \cline{2-6} 
                                  & \modelreg   & 0.83$\pm$0.25             & 0.65$\pm$0.38           & 0.99$\pm$0.37           & 1.02$\pm$0.26           \\ \cline{2-6} 
                                  & \bseqregt  & 0.62$\pm$0.19             & 1.29$\pm$0.47           & 0.92$\pm$0.27           & 1.15$\pm$0.66           \\ \cline{2-6} 
                                  & \bseqreg & 2.15$\pm$0.66             & 3.04$\pm$0.73           & 3.78$\pm$0.58           & 2.51$\pm$0.61           \\ \cline{2-6} 
                                  & \model    & \textbf{7.52$\pm$0.40}    & \textbf{7.18$\pm$0.27}  & \textbf{7.28$\pm$0.69}  & \textbf{7.59$\pm$0.88}  \\ \hline
\end{tabular}
    }
    
\end{table}
\clearpage

\end{document}